\documentclass[letterpaper]{article} 
\usepackage{aaai2027}
\usepackage[hyphens]{url}  
\usepackage{graphicx} 
\usepackage{natbib}  
\usepackage{caption} 
\usepackage{amsmath}
\usepackage{array}
\usepackage{graphicx}
\usepackage{subcaption}

\usepackage{algorithm}
\usepackage{algorithmic}

\usepackage{newfloat}
\usepackage{listings}
\DeclareCaptionStyle{ruled}{labelfont=normalfont,labelsep=colon,strut=off} 
\floatstyle{ruled}
\newfloat{listing}{tb}{lst}{}
\floatname{listing}{Listing}

\usepackage{booktabs}
\usepackage{amssymb}
\usepackage{multirow}
\usepackage{pifont}

\title{DC-WAM: Dynamic-Centric Visual Supervision and Reasoning for World-Action Models}
\author {
    Haoyuan Ji\textsuperscript{\rm 1,\rm 2}\equalcontrib,
    Lingxiang Fan\textsuperscript{\rm 1,\rm 2}\equalcontrib,
    Shang Su\textsuperscript{\rm 1,,\rm 2},
    Yinqiao Lu\textsuperscript{\rm 1},
    Mengkai Shi\textsuperscript{\rm 2},
    Jun Gao\textsuperscript{\rm 3}\corresponding,
    Shuo Feng\textsuperscript{\rm 1}\corresponding
}
\affiliations {
    \textsuperscript{\rm 1}Tsinghua University\\
    \textsuperscript{\rm 2}Dense-AI\\
    \textsuperscript{\rm 3}University of Michigan\\
}

\begin{document}

\nocopyright 
\maketitle

\begin{abstract}

World-Action Models (WAMs) augment robot policies with future visual prediction, but it remains unclear what the visual modality should learn for control. While photorealistic future prediction provides dense supervision, it also incurs substantial computation and can allocate capacity to texture, illumination, and background variations that are only weakly related to action selection. Recent efficient WAM variants suggest that the main benefit of the video branch may not lie in the rendered future itself, but in the control-relevant visual representations induced during training. In this work, we revisit future video prediction from a dynamic-centric perspective and ask whether an existing RGB-based WAM can be redirected from appearance-dominated reconstruction toward interaction-induced visual dynamics without introducing additional modality-specific predictions or online inputs at deployment. We propose DC-WAM, a dynamic-centric WAM framework that redistributes supervision and computation in the RGB video branch. At the supervision level, DC-WAM combines temporal-difference flow matching with trajectory-guided weighting, emphasizing dense temporal changes and localized regions where the gripper, manipulated objects, and contact areas move. At the reasoning level, DynaRoute predicts token-wise dynamic relevance and converts it into an attention bias, guiding the model toward control-relevant future tokens. Experiments in simulation and on real-world manipulation tasks show that DC-WAM consistently improves policy performance, especially under out-of-distribution perturbations in lighting, object appearance, and background texture.

\end{abstract}


\section{Introduction}


\begin{figure}[t]
\centering
\includegraphics[width=0.95\columnwidth]{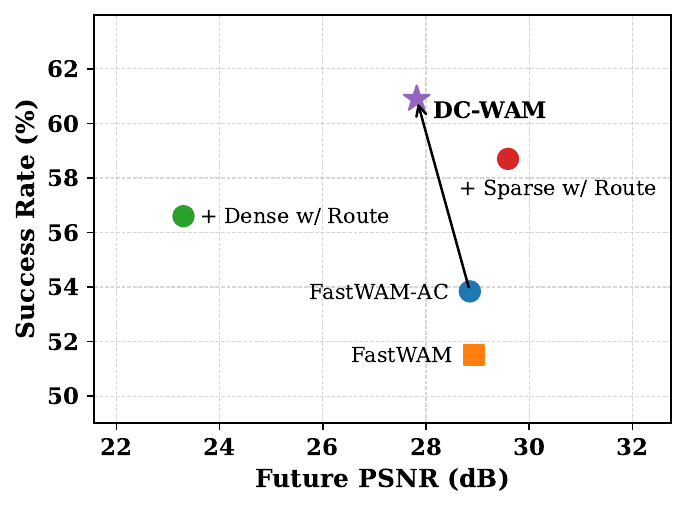}
\caption{Effect of dynamic-centric supervision and routing.
    Our method achieves a relatively low PSNR, but improves policy success on LIBERO-Plus evaluation, suggesting that appearance-level reconstruction fidelity is not aligned with control-relevant future prediction.
    }
\label{fig:sr_psnr_analysis}
\end{figure}

Vision-Language-Action (VLA) models map visual observations and language instructions directly to robot actions, leveraging the semantic priors of large vision-language backbones~\cite{openvla,pi0}. World-Action Models (WAMs) further couple action generation with future visual prediction, using anticipated scene evolution as an additional learning signal for robot control~\cite{uwm,uva,dreamzero,lingbotva,kim2026cosmospolicy}. By jointly modeling what the robot should do and how the scene may change, WAMs provide dense temporal supervision beyond action imitation alone.

However, uniform RGB future prediction entangles manipulation dynamics with appearance factors such as texture, illumination, background, and sensor noise. Although these factors dominate reconstruction errors, they often change across environments without altering the underlying manipulation dynamics, making appearance-centric supervision potentially brittle under visual distribution shifts. Consequently, an RGB video branch may allocate substantial capacity to appearance reconstruction rather than the gripper motion, object displacement, and contact events that are more directly useful for action learning.

Recent efficient WAMs show that strong action policies do not necessarily require complete future videos to be iteratively generated during execution~\cite{fastwam,efficientwam,imagewam}. Meanwhile, other methods reduce the ambiguity of RGB prediction by introducing structured future representations, such as semantic masks, point trajectories, optical flow, or geometric states~\cite{MaskWAM,JOPAT,fofpred,oawam,MECo-WAM}. Although these representations provide more structured supervision, they typically modify or expand the predicted future-state modality. We instead ask whether the existing RGB-based WAM can be redirected toward dynamic-centric representations that remain stable across appearance shifts, without additional modality-specific prediction.

We propose \textbf{DC-WAM}, a dynamic-centric World-Action Model that redistributes the supervision and reasoning toward interaction-induced dynamics. At the supervision level, DC-WAM replaces uniform latent reconstruction with two complementary training signals. First, temporal-difference supervision emphasizes changes between adjacent visual flow fields, reducing the influence of temporally persistent appearance components. Second, tracker-guided flow matching reweights the visual objective toward localized regions with strong gripper, object, and contact motion. The tracker-derived targets are constructed offline from training videos and are not required during policy execution. At the reasoning level, we introduce \textbf{DynaRoute}, a lightweight module that predicts token-wise dynamic relevance and converts it into an attention bias. This routes visual attention toward future tokens associated with interaction-induced motion. 

We retain an action-conditioned visual branch, allowing video supervision to regularize action representations through an explicit action-to-visual pathway~\cite{ye2026gigaworldpolicy}. The RGB branch is used during training but can be removed at deployment following Fast-WAM-style inference, preserving efficient action generation.


Experiments show that DC-WAM substantially improves robustness under distribution shift. As summarized in Fig.~\ref{fig:sr_psnr_analysis}, the proposed method progressively improves LIBERO-Plus success, demonstrating that DC-WAM improves OOD generalization while maintaining strong clean-condition performance.

In summary, our contributions are threefold:

\begin{itemize}
    \item We propose DC-WAM, which redirects the existing RGB video branch from appearance-dominated reconstruction toward interaction-induced visual dynamics, without introducing additional modality-specific prediction branches or deployment-time inputs.

    \item We introduce complementary dynamic-centric supervision and reasoning mechanisms: temporal-difference supervision suppresses persistent appearance components, tracker-guided flow matching emphasizes localized manipulation dynamics, and DynaRoute routes
    visual attention toward dynamically relevant tokens.

    \item Trained only on clean demonstrations, DC-WAM maintains strong in-distribution performance while substantially improving OOD success and reducing ID--OOD degradation on LIBERO-Plus and real-world manipulation under unseen visual perturbations.
\end{itemize}

\section{Related Work}
\label{sec:related}



\subsection{World-Action Models and Efficient Visual Foresight}

Recent WAMs couple video prediction with action generation through modality-specific diffusion, shared visual-action representations, or interleaved generation, achieving strong manipulation performance \cite{uwm,uva,dreamzero,lingbotva,ye2026gigaworldpolicy}. However, generating complete RGB futures introduces substantial inference cost. Recent methods reduce this dependence by removing the video branch at deployment, predicting coarse visual futures with compact experts, or using intermediate features from image-editing backbones~\cite{vpp,fastwam,efficientwam,imagewam}. These results suggest that the value of visual foresight may lie more in the predictive representations learned during training than in the rendered future itself.

\subsection{Structured Visual Representations for Control}

Prior VLA methods reduce the spatial ambiguity of current visual observations using trajectory sketches, state-action traces, or image-space target prompts \cite{rttrajectory,tracevla,aimbot}. World-model policies further introduce structured visual representations into future prediction by replacing or augmenting RGB futures with semantic masks, point trajectories, optical flow, object states, or geometric representations \cite{maskworldmodel,MaskWAM,JOPAT,fofpred,oawam,MECo-WAM}. These representations make future prediction more structured and action-relevant, but typically modify the predicted future representation or introduce additional modality-specific tokens. 
DC-WAM instead retains the RGB-based visual branch and uses dynamic cues only to reweight visual supervision and attention routing. Our use of trajectory-derived supervision is related to a physically grounded video generation method PhysisForcing \cite{zhang2026physisforcing}, but DC-WAM applies these cues inside a visual-action WAM through temporally resolved token-level relevance maps coupled with action generation.

\subsection{Attention Bias for Action Reasoning}



Attention-logit biases inject structured priors without modifying the token set~\cite{shaw2018selfattn,press2022ALiBi}. Unlike static positional biases, DynaRoute adopts a similar logit-level modulation mechanism, but replaces static positional priors with input-dependent dynamic relevance. 
It predicts a relevance score for each future visual token and converts it into a key-side bias in visual-token attention, emphasizing regions associated with gripper motion, object displacement, and contact changes.

\begin{figure*}[t]
    \centering
    \includegraphics[width=\textwidth]{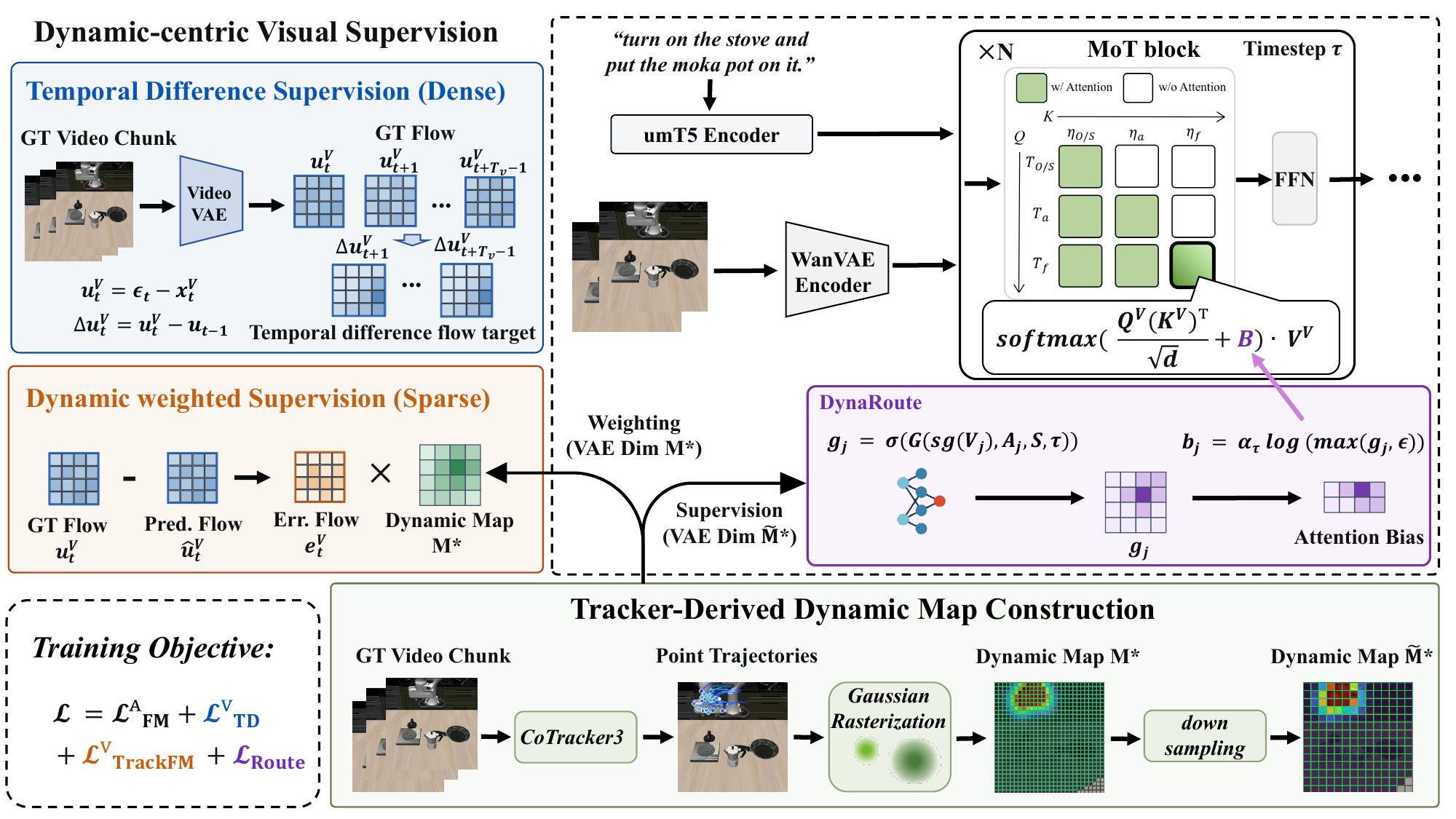}
    \caption{
    Overview of DC-WAM. DC-WAM redirects the RGB video branch toward interaction-induced dynamics through complementary dense and sparse supervision. Temporal-difference supervision models changes in adjacent visual flows, while TrackFM reweights visual flow-matching errors using an offline tracker-derived dynamic map. The same map supervises DynaRoute, which predicts token-wise dynamic relevance and converts it into a key-side bias in the visual-action MoT attention. Tracker-derived targets are used only during training and require no additional online input or modality-specific predictions at deployment.
    }
    \label{fig:dcwam_overview}
\end{figure*}

\providecommand{\CaseXLIVRoot}
{images/case_044__libero_spatial__b04__noise}

\newcommand{\AttnImage}[1]{%
    \raisebox{-0.5\height}{%
        \includegraphics[width=0.235\textwidth]{#1}%
    }%
}

\newcommand{\AttnLabel}[1]{%
    \makebox[1.25em][c]{%
        \smash{%
            \raisebox{-0.5\height}{%
                \rotatebox[origin=c]{90}{%
                    \small\bfseries\strut #1%
                }%
            }%
        }%
    }%
}

\begin{figure*}[t]
    \centering
    \begingroup
    \setlength{\tabcolsep}{0.5pt}

    \begin{tabular}{@{}c@{\hspace{0.8pt}}cccc@{}}
        &
        \textbf{Clean}
        &
        \textbf{L1}
        &
        \textbf{L2}
        &
        \textbf{L3}
        \\[0.8pt]

        \AttnLabel{Image}
        &
        \AttnImage{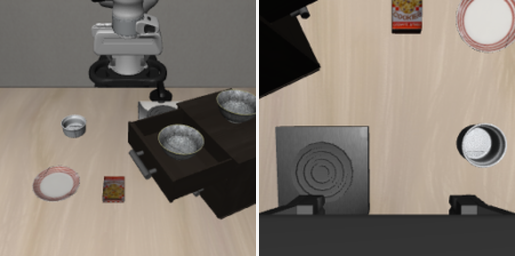}
        &
        \AttnImage{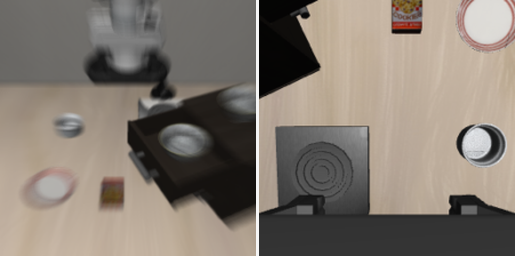}
        &
        \AttnImage{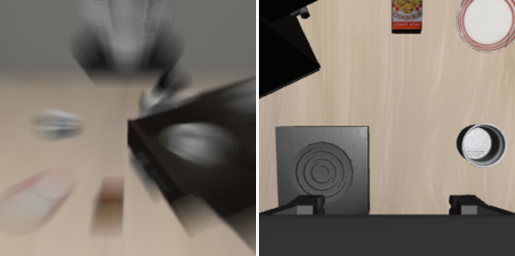}
        &
        \AttnImage{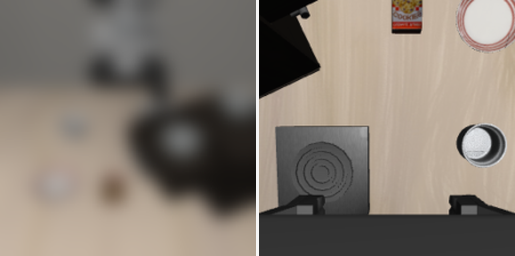}
        \\[0.6pt]

        \AttnLabel{FastWAM-AC}
        &
        \AttnImage{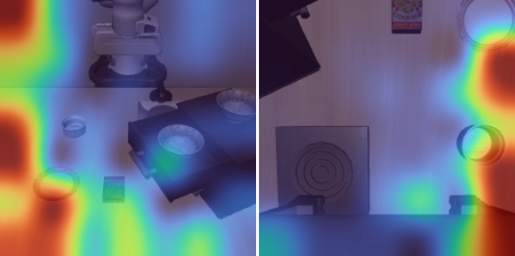}
        &
        \AttnImage{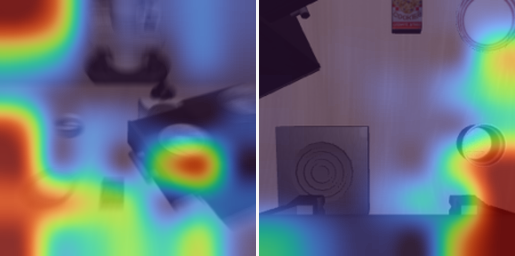}
        &
        \AttnImage{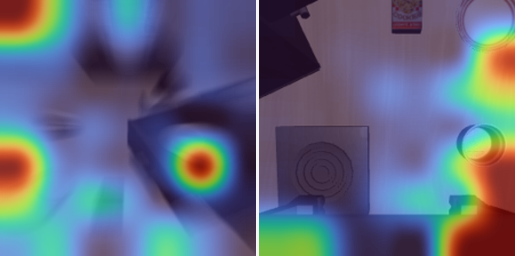}
        &
        \AttnImage{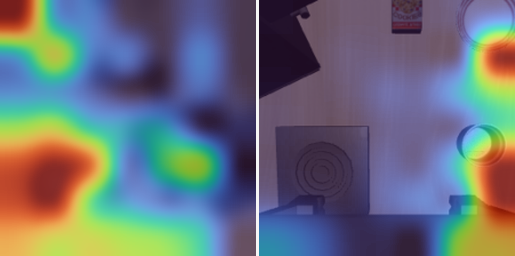}
        \\[0.6pt]

        \AttnLabel{DC-WAM}
        &
        \AttnImage{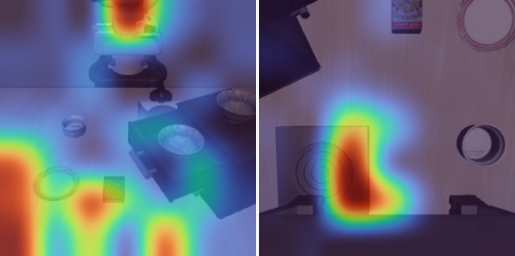}
        &
        \AttnImage{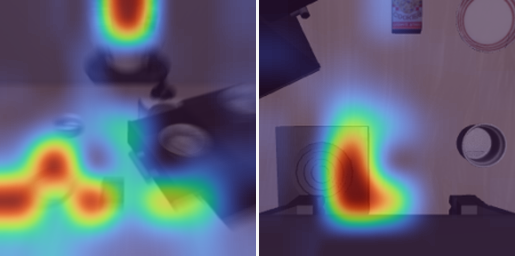}
        &
        \AttnImage{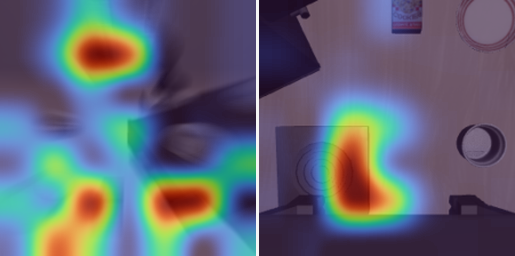}
        &
        \AttnImage{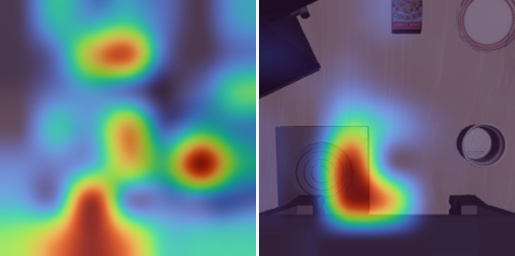}
    \end{tabular}%
    
    \endgroup

    \caption{
    Attention responses under increasing visual perturbations.
    As perturbation severity increases, FastWAM-AC attention becomes
    increasingly diffuse and shifts toward static distractors, whereas
    DC-WAM maintains more consistent and concentrated attention on
    manipulation-relevant dynamic regions, indicating stronger attention
    stability and OOD robustness under visual corruption.
    }
    \label{fig:case044-clean-noise-3x4}
\end{figure*}

\section{Method}
\label{sec:method}

\paragraph{Overview.}

We consider language-conditioned robotic manipulation from visual observations. Given the current RGB image $\mathbf{o}_t \in \mathbb{R}^{3\times H\times W}$, language instruction $\ell$, and proprioceptive state $\mathbf{s}_t$, a World-Action Model (WAM) jointly predicts an action chunk and future visual evolution:
\begin{equation}
    p_{\theta}\!\left(
    \mathbf{a}_{t:t+T_a-1},
    \mathbf{\hat{o}}_{t+1:t+T_v}
    \mid
    \mathbf{o}_t, \ell, \mathbf{s}_t
    \right),
\label{eq:wam_distribution}
\end{equation}
where $T_a$ and $T_v$ denote the action and visual prediction horizons, respectively.

\paragraph{Architecture.}\label{sec:architecture}

DC-WAM is built on the Wan2.2 backbone~\cite{wan2025} and follows a visual-action MoT architecture with two branches: an RGB visual branch for future latent video prediction and an action branch for action-chunk generation. We keep the original RGB-based visual branch and do not introduce additional modality-specific experts as extra token vectors in MoT computation.

The visual branch is action-conditioned through the MoT attention mask. Future visual tokens can attend to action tokens, while action tokens are prevented from attending to future visual tokens and can only attend to the current observed visual tokens, avoiding future leakage during action prediction.

Building on this backbone, DC-WAM modifies the training and forward computation of the RGB visual branch in two ways: dynamic-centric visual supervision and DynaRoute attention bias, as illustrated in Fig.~\ref{fig:dcwam_overview}. 
We describe these two components next.

\subsection{Tracker-Derived Dynamic Map Construction}
\label{sec:dynamic_map}

We construct tracker-derived dynamic maps offline at the episode level, providing concentration on manipulation dynamics. They are not used during policy execution.

Given an episode with $T_{\mathrm{ep}}$ frames, we sample candidate points either uniformly over the image or within foreground regions produced by SAM~\cite{SAM}. An off-the-shelf point tracker~\cite{cotracker3} estimates their trajectories throughout the episode. Let the position of the $n$-th tracked point at global time $t$ be
\begin{equation}
    \mathbf{p}_{t,n} = (x_{t,n}, y_{t,n}) \in [0,W) \times [0,H),
\end{equation}
where $t=0,\ldots,T_{\mathrm{ep}}-1$. We compute the frame-wise motion magnitude using the backward temporal difference:
\begin{equation}
    d_{t,n} = \left\|\mathbf{p}_{t,n} - \mathbf{p}_{t-1,n}
    \right\|_2,\quad t=1,\ldots,T_{\mathrm{ep}}-1,
\label{eq:point_displacement}
\end{equation}
and set $d_{0,n}=0$. 
Dynamic points at time $t$ are selected by
\begin{equation}
    \mathcal{P}^{t}_{\mathrm{dyn}}=\left\{n \mid d_{t,n} > \delta_{\mathrm{mot}}\right\},
\label{eq:dynamic_point_set}
\end{equation}
where $\delta_{\mathrm{mot}}$ filters out static points and small tracking fluctuations.

We rasterize the selected dynamic points onto the VAE visual-token grid. Let $(x_k,y_k)$ denote the center of the $k$-th visual token in the original image coordinate system, where $k=1,\ldots,H_zW_z$ and $H_z\times W_z$ is the spatial resolution of the VAE visual latent. The spatial response between point $n$ and token $k$ is
\begin{equation}
    \kappa_{t,n,k}=\exp\left[-\lambda\left(
    \frac{(x_k-x_{t,n})^2}{\sigma_x^2} +
    \frac{(y_k-y_{t,n})^2}{\sigma_y^2}\right)\right],
\label{eq:dynamic_kernel}
\end{equation}
with
$
    \sigma_x = \frac{W}{W_z}\sigma_p, \quad \sigma_y = \frac{H}{H_z}\sigma_p.
$
Unless otherwise specified, we set $\lambda=0.25$ and $\sigma_p=1.25$.

The unnormalized token-level dynamic response is obtained by aggregating motion-weighted kernel responses:
\begin{equation}
    b_{t,k} = \sum_{n\in \mathcal{P}^{t}_{\mathrm{dyn}}}
    d_{t,n}\kappa_{t,n,k}.
\label{eq:dynamic_response}
\end{equation}
Finally, we normalize the response over the entire episode and within each camera view:
\begin{equation}
    m^{*}_{t,k} = \frac{b_{t,k}}{\max_{t',k'} b_{t',k'} + \epsilon} \in [0,1].
\label{eq:dynamic_map}
\end{equation}

This episode-level normalization preserves both spatial and temporal saliency: it highlights tokens near strong tracked motion and assigns larger relevance values to time steps where interaction-induced changes are more pronounced. 

An important consequence of this construction is its reduced
sensitivity to temporally persistent appearance shifts.
Because the dynamic map is derived from inter-frame point
displacements rather than RGB reconstruction errors, static or
slowly varying changes in illumination and background texture
do not directly contribute to the supervision target, provided
that the underlying point trajectories remain stable.
Consequently, the visual branch is encouraged to prioritize
gripper motion, object displacement, and contact-related changes
instead of fitting nuisance appearance variations. This provides
a natural source of robustness to appearance-level OOD
perturbations, such as lighting and background changes, that alter
visual appearance without changing the underlying manipulation
dynamics.

These maps are used to reweight the original visual flow-matching loss toward sparse interaction regions and to supervise the dynamic relevance predicted by DynaRoute, whose outputs are defined on the DiT visual-token grid. We further downsample the map to the DiT token resolution $H_D \times W_D$:
\begin{equation}
    \widetilde{m}^{*}_{t,k}
    =
    \mathcal{D}_{\mathrm{tok}}
    \left(
    m^{*}_{t,\cdot}
    \right)_k,
    \quad
    k=1,\ldots,H_DW_D,
\label{eq:token_dynamic_map}
\end{equation}
where $\mathcal{D}_{\mathrm{tok}}$ denotes patch-wise downsampling from the VAE latent grid to the DiT token grid. 

\subsection{Dynamics-Aware Attention Bias}
\label{sec:dynaroute}

DynaRoute predicts token-wise relevance for future visual tokens and
converts it into an additive key-side bias in the visual branch.
It encourages the video expert to prioritize interaction-induced
dynamics, such as object displacement and robot-environment contact,
rather than attending uniformly to all visual tokens.

Let
$\mathbf V_{\tau}\in\mathbb R^{B\times S_v\times d_v}$
denote the DiT visual-token sequence, where
$S_v=T_vK_D$, $K_D=H_DW_D$, and $d_v$ is the visual hidden
dimension. We decompose it as
\begin{equation}
    \mathbf V_{\tau}
    =
    [\mathbf V^{\mathrm{obs}},
     \mathbf V_{\tau}^{\mathrm{fut}}],
\end{equation}
where $\mathbf{V}^{\mathrm{obs}}$ contains the clean current-observation tokens, while $\mathbf{V}^{\mathrm{fut}}_{\tau}$ represents the
future visual tokens at timestep $\tau$. The observation tokens remain clean at all timesteps. Let $\mathbf A_{\tau}$ denote the action tokens at the same timestep, and let $\mathbf S$ denote the fused language and proprioceptive conditioning tokens.

DynaRoute is evaluated once at each diffusion timestep:
\begin{equation}
    \mathbf z_{\tau}
    =
    G_{\psi}\!\left(
        \operatorname{sg}(\mathbf V_{\tau}),
        \operatorname{sg}(\mathbf A_{\tau}),
        \mathbf S,
        \tau
    \right)
    \in \mathbb R^{B\times S_v},
\label{eq:dynaroute_logits}
\end{equation}
where $\operatorname{sg}(\cdot)$ denotes stop-gradient. The predicted dynamic relevance is
\begin{equation}
    \mathbf g_{\tau}
    =
    \sigma(\mathbf z_{\tau})
    \in (0,1)^{B\times S_v},
\label{eq:dynamic_relevance_prediction}
\end{equation}
and is supervised by the downsampled tracker-derived map $\widetilde{\mathbf m}^{*}$ defined in Eq.~\ref{eq:token_dynamic_map}.

For each future visual token $i$, we convert its relevance into a log-space bias:
\begin{equation}
    b_i
    =
    \alpha
    \log\!\left(\max(g_i,\epsilon)\right),
\label{eq:dynaroute_bias}
\end{equation}
where $\alpha$ controls the routing strength and $\epsilon$ ensures numerical stability. Low-relevance tokens therefore receive stronger negative biases. We further center the bias over visual tokens to stabilize the bias scale:
\begin{equation}
    \bar b_i = b_i- \frac{1}{S_v}\sum_{q=1}^{S_v}b_q.
\label{eq:centered_dynaroute_bias}
\end{equation}

Let $\mathbf Q_\ell^V$, $\mathbf K_\ell^V$, and $\mathbf V_\ell^V$ denote the visual queries, keys, and values at the $\ell$-th MoT layer. The same bias is shared across all layers at the current diffusion timestep:
\begin{equation}
    \operatorname{Attn}_{\ell}^{V\leftarrow V}
    =
    \operatorname{softmax}\!\left(
        \frac{\mathbf Q_\ell^V(\mathbf K_\ell^V)^\top}{\sqrt d} +
        \bar{\mathbf B}\right)\mathbf V_\ell^V,
\label{eq:dynaroute_attention}
\end{equation}
where $\bar{\mathbf B}$ is obtained by broadcasting $\bar{\mathbf b}$ over attention heads and visual-query positions. 

Additional implementation details of DynaRoute, together with further analyses of its routing behavior and effectiveness, are provided in Appendix.

\paragraph{Action-only inference with routed video cache.}

\begin{figure}[t]
\centering
\includegraphics[width=0.95\columnwidth]{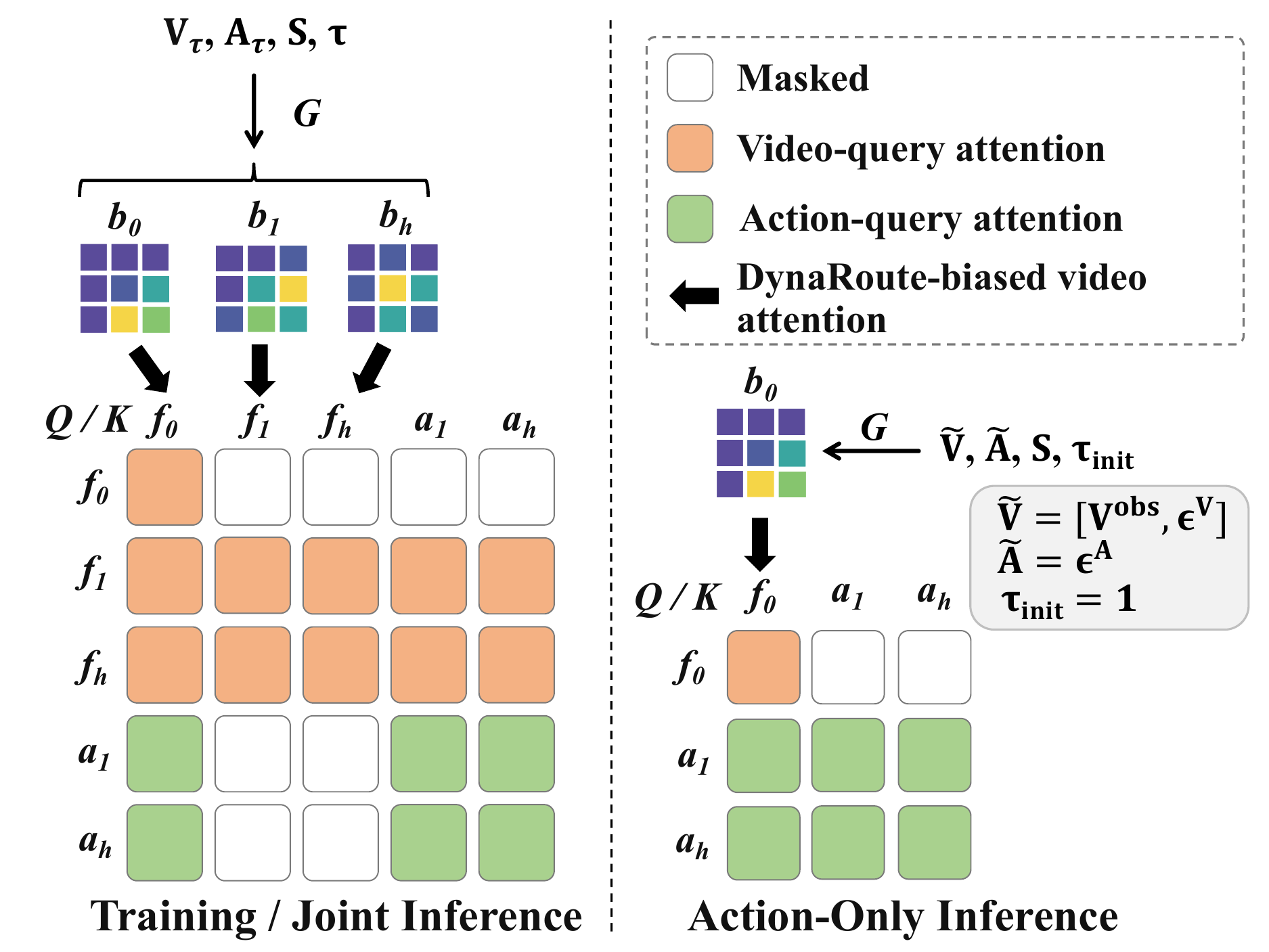}
\caption{DynaRoute during training and inference. During training, a layer-shared bias is predicted at each diffusion timestep and injected into video-query attention.
At deployment, DynaRoute is evaluated once at $\tau_{\mathrm{init}}=1$ to construct the routed visual cache $\mathcal C_V$. The video branch is then disabled and the cache is
reused throughout action denoising.
}
\label{fig:action_only_inference}
\end{figure}

As illustrated in Fig.~\ref{fig:action_only_inference}, DC-WAM follows
Fast-WAM-style action-only inference at deployment. Rather than
iteratively denoising future video, it executes the video branch only
once to construct a routed visual key-value cache $\mathcal C_V$ for
the action branch.

At the initial diffusion step $\tau_{\mathrm{init}}=1$, we form the
pseudo-video and noisy action inputs as
\begin{equation}
    \widetilde{\mathbf V}
    =
    [\mathbf{V}^{\mathrm{obs}},\boldsymbol{\epsilon}^{V}],
    \quad
    \boldsymbol{\epsilon}^{V}
    \sim\mathcal N(\mathbf 0,\mathbf I),
\end{equation}
and
\begin{equation}
    \widetilde{\mathbf A}
    =
    \boldsymbol{\epsilon}^{A},
    \quad
    \boldsymbol{\epsilon}^{A}
    \sim\mathcal N(\mathbf 0,\mathbf I),
\end{equation}
where $\mathbf{V}^{\mathrm{obs}}$ is the clean observed-frame latent and $\boldsymbol{\epsilon}^{V}$ occupies the future visual slots. DynaRoute is evaluated once to produce
\begin{equation}
    \bar{\mathbf b}_{\mathrm{cache}}
    =
    \mathrm{Bias}\!\left(
    G_{\psi}\!\left(
    \operatorname{sg}(\widetilde{\mathbf V}),
    \operatorname{sg}(\widetilde{\mathbf A}),
    \mathbf S,
    \tau_{\mathrm{init}}
    \right)\right),
\end{equation}
which is injected during video-cache prefill. After constructing
$\mathcal C_V$, the video branch is no longer executed and the action
branch reuses the cached visual keys and values for all subsequent
denoising steps.

This design preserves train-inference consistency for DynaRoute, as it closely matches the high-noise regime used during training. At large diffusion timesteps, DynaRoute infers dynamic relevance primarily from the clean observation, language instruction, and proprioceptive state; at lower-noise timesteps, it can additionally exploit partially preserved information in the action and future-visual tokens.

\subsection{Training Objective}
\label{sec:training_objective}

The complete training objective is
\begin{equation} 
    \mathcal{L} = \mathcal{L}_{\mathrm{FM}}^{A} + \mathcal{L}_{\mathrm{TD}}^{V} + \mathcal{L}_{\mathrm{TrackFM}}^{V} + \mathcal{L}_{\mathrm{Route}}, 
\label{eq:total_loss}
\end{equation}
where the four terms supervise action generation, dense temporal dynamics, sparse interaction regions, and DynaRoute relevance prediction, respectively.

\paragraph{Flow matching.}
For a clean sample $\mathbf{x}$ and Gaussian noise $\boldsymbol{\epsilon}\sim\mathcal{N}(\mathbf{0},\mathbf{I})$, we use the linear interpolation
\begin{equation}
    \mathbf{x}_{\tau}=(1-\tau)\mathbf{x}+  \tau\boldsymbol{\epsilon}, \quad \tau\sim\mathcal{U}(0,1),
\label{eq:fm_interpolation}
\end{equation}
with the flow target
\begin{equation}
    \mathbf{u}
    =
    \boldsymbol{\epsilon}-\mathbf{x}.
\label{eq:fm_velocity}
\end{equation}
For actions, the model predicts $\widehat{\mathbf{u}}^{A}$ and is trained with
\begin{equation}
    \mathcal{L}_{\mathrm{FM}}^{A}
    =
    \mathbb{E}
    \left[
    \left\|
    \widehat{\mathbf{u}}^{A}
    -
    \mathbf{u}^{A}
    \right\|_2^2
    \right].
\label{eq:action_fm}
\end{equation}

To emphasize state transitions, we impose a temporal-difference loss on the visual flow velocity at the VAE latent resolution \cite{DreamDojo}. Let $\widehat{\mathbf{u}}^{V}_t$ and $\mathbf{u}^{V_t}=\boldsymbol{\epsilon}^{V}_t-\mathbf{x}^{V}_t$ denote the predicted visual velocity and the target flow velocity at time step t. The temporal difference objective $\mathcal{L}_{\mathrm{TD}}^{V}$ enforces temporal consistency by matching adjacent-frame transitions between the predicted and target visual velocity fields:
\begin{equation}
    \mathcal{L}_{\mathrm{TD}}^{V} = \mathbb{E}\Bigg[
    \sum_{t=1}^{T_v-1}
    \Big\|(\widehat{\mathbf{u_t}}^{V}-\widehat{\mathbf{u}}^{V}_{t-1}) - (\mathbf{u}^{V}_{t}-\mathbf{u}^{V}_{t-1})\Big\|_2^2\Bigg].
\label{eq:temporal_loss}
\end{equation}

This temporal-difference loss suppresses temporally invariant appearance components, but it does not explicitly localize task-critical interaction regions. 
We therefore use the tracker-derived dynamic map $\mathbf{m}^{*}\in[0,1]^{T_v\times H_zW_z}$ to reweight the original visual FM error on the VAE latent grid. 
Since $\mathbf{m}^{*}$ is constructed from thresholded tracked motion, most static background cells have zero or near-zero weights, making the map spatially sparse. 
For each latent frame $t$ and cell $p$, let
\begin{equation}
    e^{V}_{t,p} =
    \left\|\widehat{\mathbf{u}}^{V}_{t,p} - \mathbf{u}^{V}_{t,p}\right\|_2^2 .
\label{eq:visual_fm_error}
\end{equation}

The tracker-guided objective is
\begin{equation}
    \mathcal{L}_{\mathrm{TrackFM}}^{V}
    =
    \mathbb{E}
    \left[
    \frac{
    \sum_{t,p} m^{*}_{t,p} e^{V}_{t,p}
    }{
    \sum_{t,p} m^{*}_{t,p} + \epsilon
    }
    \right].
\label{eq:track_fm}
\end{equation}

Thus, TrackFM redistributes the visual FM loss toward sparse regions with strong tracked motion, such as the end effector, manipulated objects, and contact areas. 
Together, the temporal-difference and TrackFM objectives provide complementary dense and sparse dynamic supervision.

\paragraph{DynaRoute Relevance Prediction.}

The downsampled token-level target $\widetilde{\mathbf{m}}^{*}\in[0,1]^{T_v\times K_D}$ from Eq.~\eqref{eq:token_dynamic_map} is used as the supervision target for DynaRoute. Given the predicted relevance $\mathbf{g}\in[0,1]^{T_v\times K_D}$, we combine binary cross-entropy with soft Dice losses:
\begin{equation}
    \mathcal{L}_{\mathrm{Route}} =
    \mathcal{L}_{\mathrm{BCE}}\left(\mathbf{g},\widetilde{\mathbf{m}}^{*}\right) +
    \lambda_{\mathrm{Dice}} \mathcal{L}_{\mathrm{Dice}}    \left(\mathbf{g},\widetilde{\mathbf{m}}^{*}\right),
\label{eq:gate_loss}
\end{equation}
where
\begin{equation}
    \mathcal{L}_{\mathrm{Dice}}
    =
    1
    -
    \frac{
    2\sum_{t,k} g_{t,k}\widetilde{m}^{*}_{t,k}
    +
    \epsilon
    }{
    \sum_{t,k} g_{t,k}
    +
    \sum_{t,k} \widetilde{m}^{*}_{t,k}
    +
    \epsilon
    }.
\label{eq:dice_loss}
\end{equation}
The BCE term provides token-wise relevance supervision, while the Dice term mitigates the imbalance caused by sparse dynamic regions.

\section{Experiments}
\label{sec:experiments}

\subsection{Experimental Setup}
\label{sec:experimental_setup}

\paragraph{Baselines.}

We use FastWAM and FastWAM-AC as matched baselines under the same Wan2.2 backbone, demonstrations, action space, and evaluation protocol. FastWAM-AC retains the action-conditioned visual interaction used by DC-WAM but removes the dynamic-centric objectives and DynaRoute, providing direct controlled comparison. All models are trained on eight NVIDIA A100 GPUs.

\paragraph{LIBERO and LIBERO-Plus.}

We evaluate in-distribution policy performance on the four standard LIBERO suites: Spatial, Object, Goal, and Long~\cite{liu2023libero}, using 50 rollouts per task. We further use LIBERO-Plus \cite{fei2026liberoplus} specifically as an out-of-distribution benchmark. It perturbs the original LIBERO tasks along seven dimensions: object layout, camera viewpoint, robot initial state, language instruction, lighting, background texture, and sensor noise. Each perturbation is organized into five difficulty levels, from L1 to L5. All methods are trained on the same 2000 clean demonstrations from the standard LIBERO benchmark and evaluated with identical seeds and episode horizons.

\paragraph{Real-world Evaluation.}

\begin{figure}[t]
    \centering
    \setlength{\tabcolsep}{1pt}

    \begin{tabular}{@{}c c c c@{}}
        & \textbf{T1} & \textbf{T2} & \textbf{T3} \\

        \rotatebox{90}{\textbf{Clean}}
        & \includegraphics[width=0.30\columnwidth]{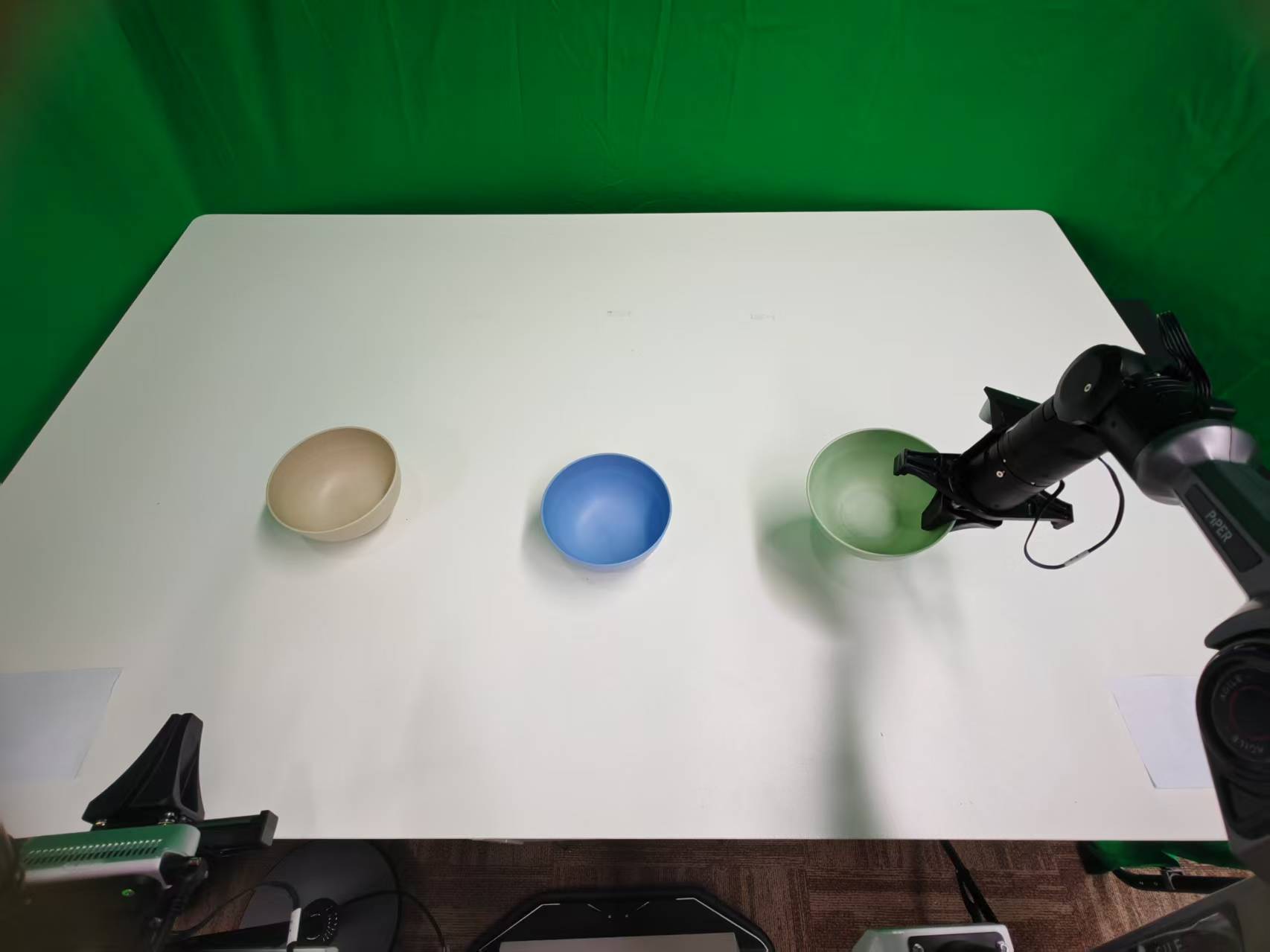}
        & \includegraphics[width=0.30\columnwidth]{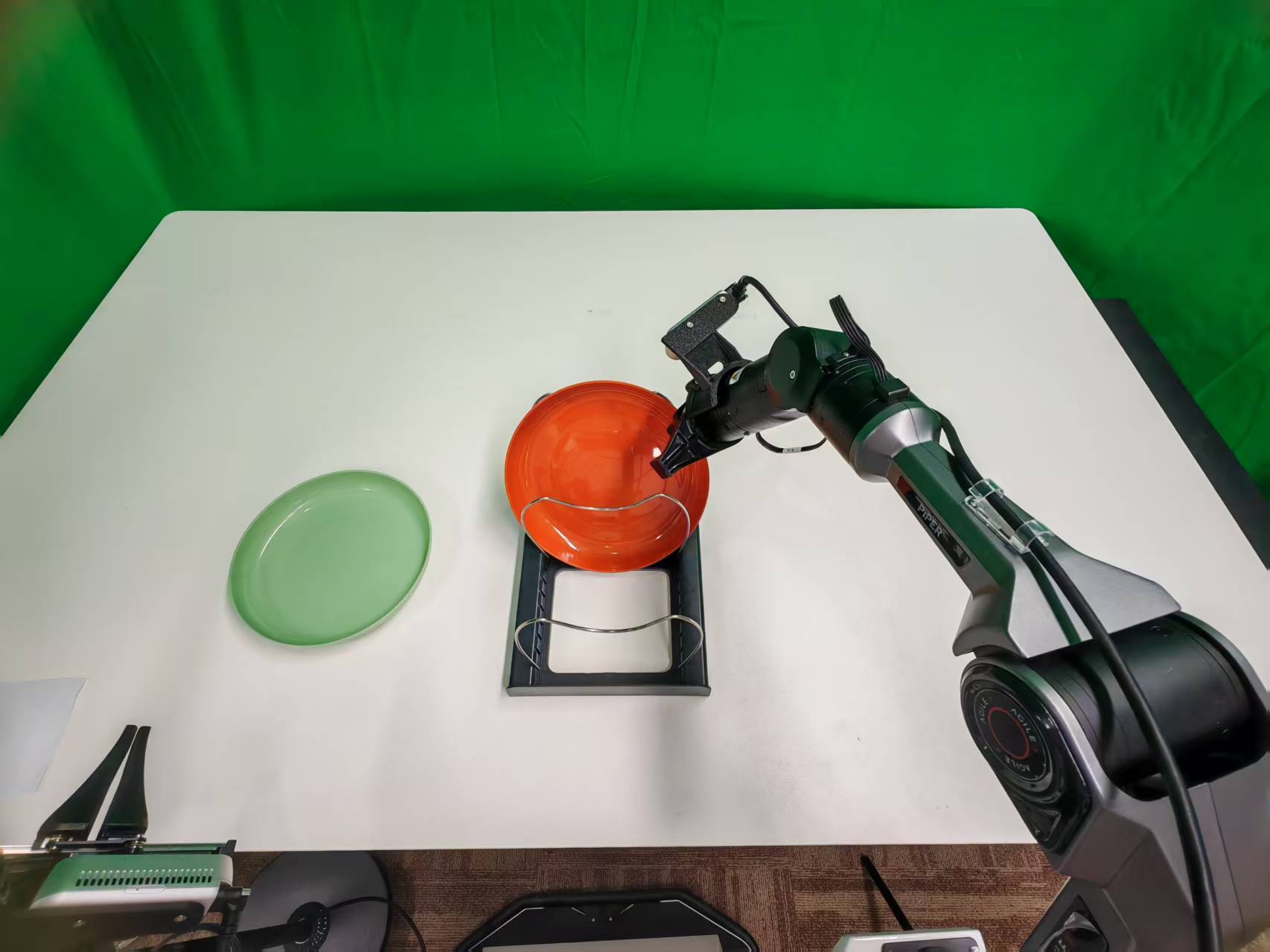}
        & \includegraphics[width=0.30\columnwidth]{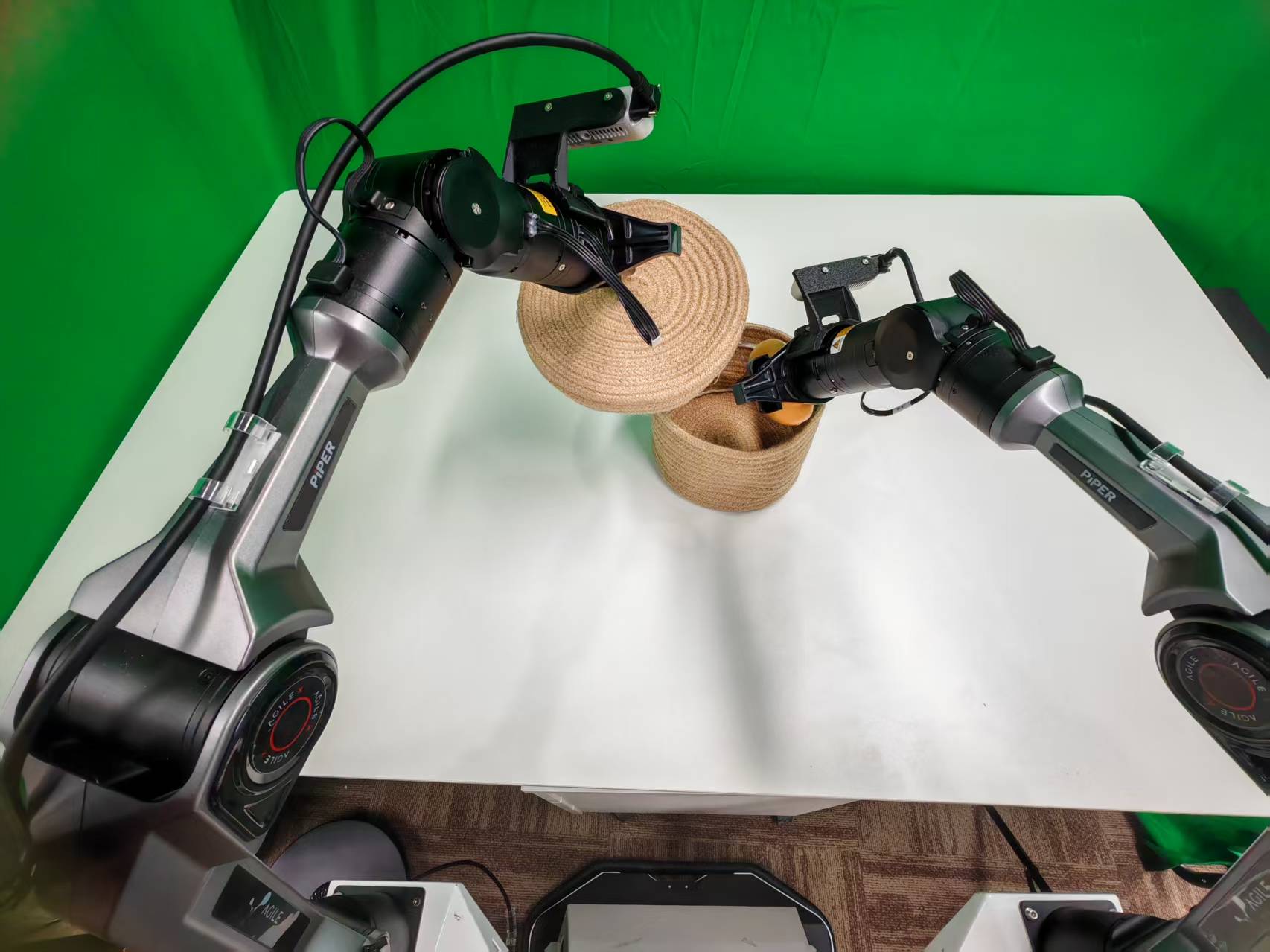} \\

        \rotatebox{90}{\textbf{Light}}
        & \includegraphics[width=0.30\columnwidth]{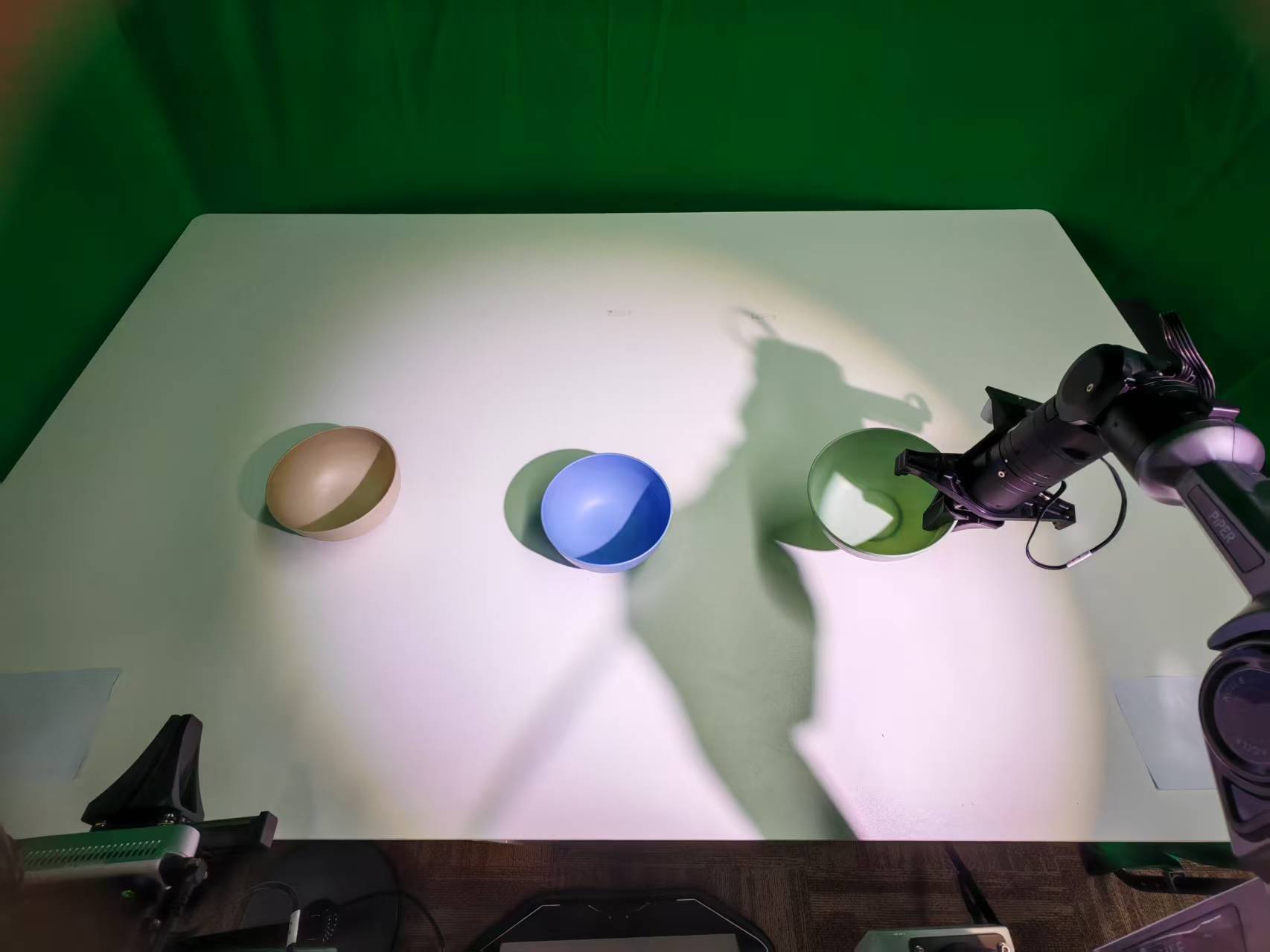}
        & \includegraphics[width=0.30\columnwidth]{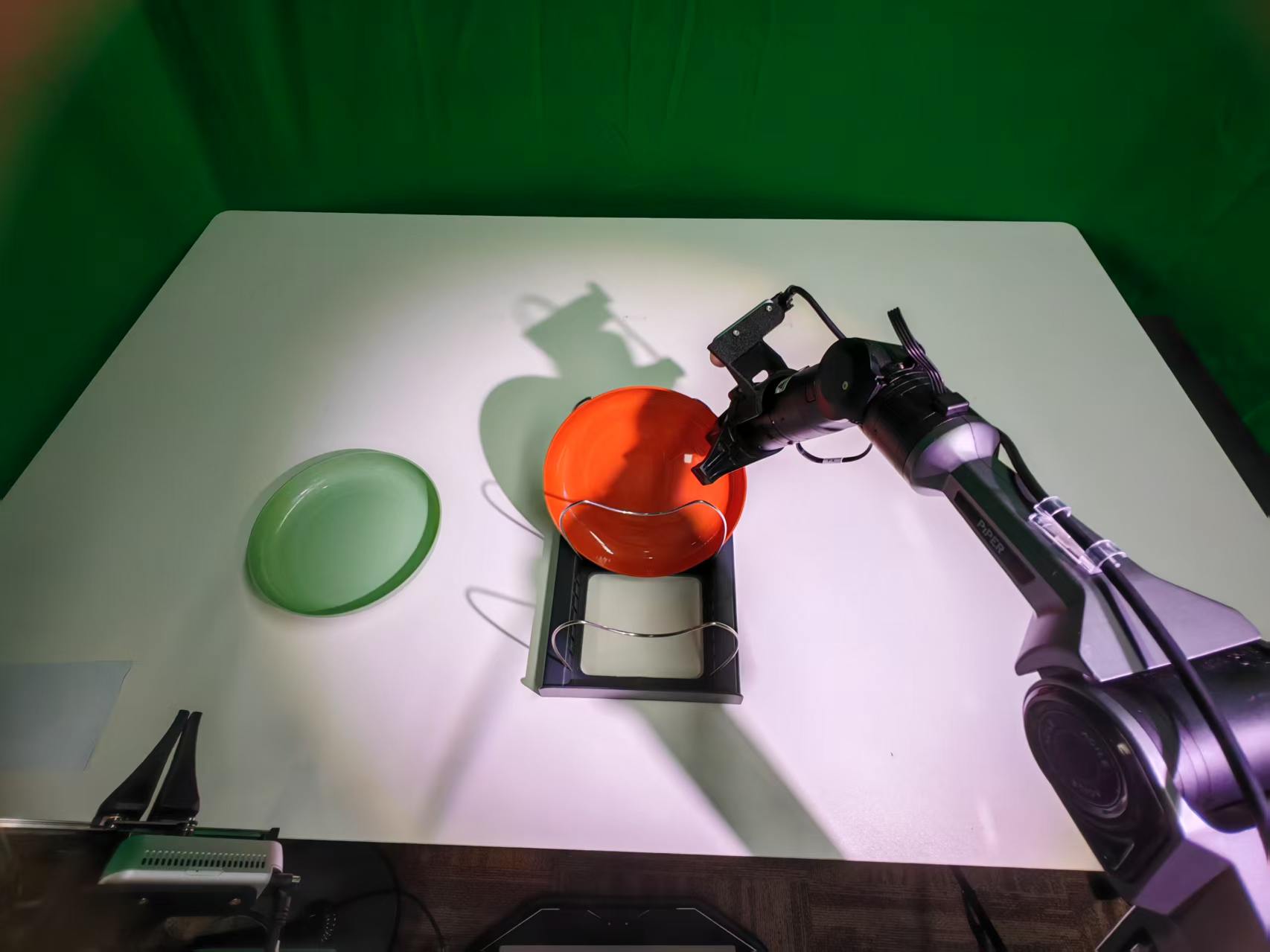}
        & \includegraphics[width=0.30\columnwidth]{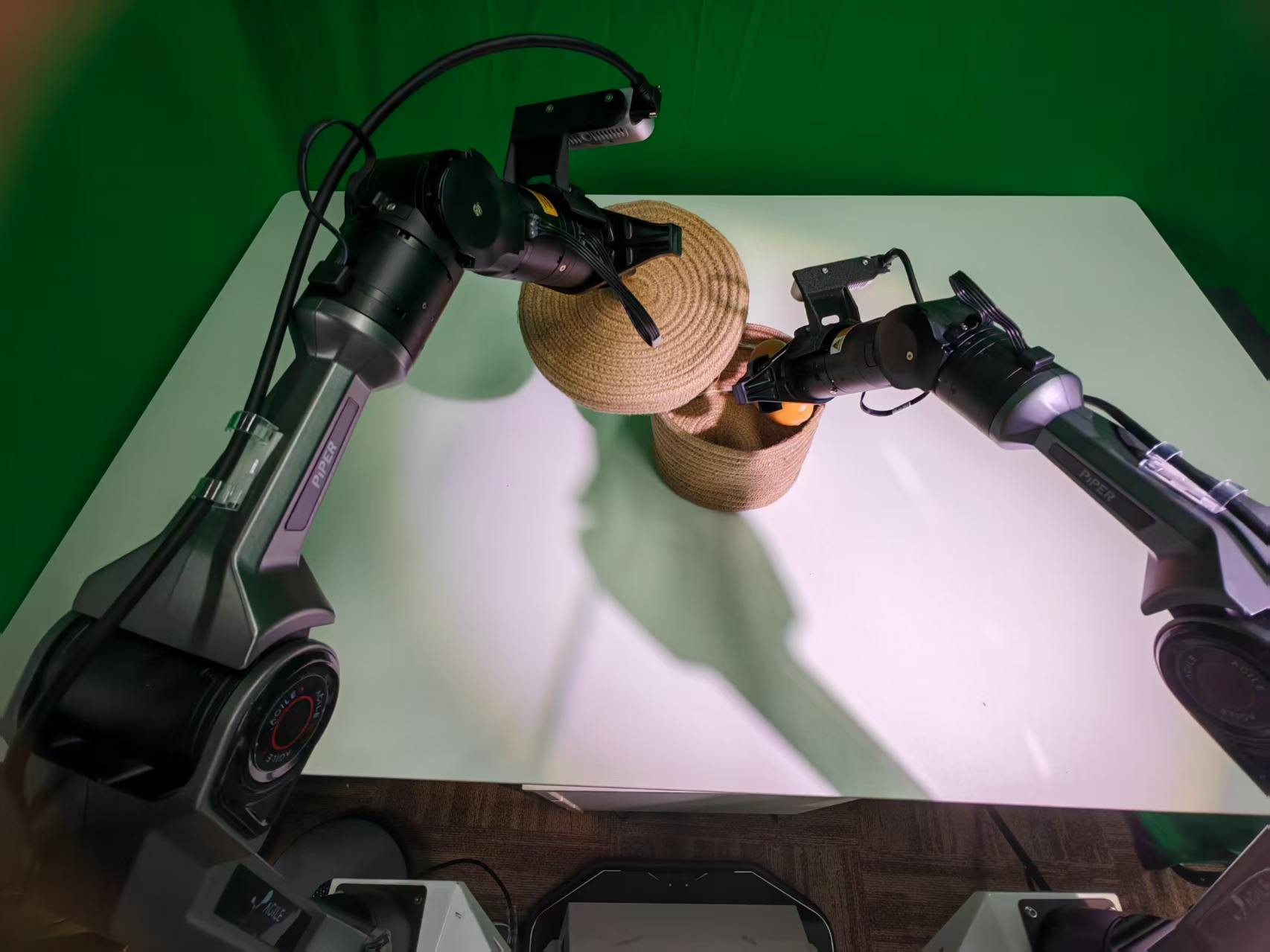} \\

        \rotatebox{90}{\textbf{Bg.}}
        & \includegraphics[width=0.30\columnwidth]{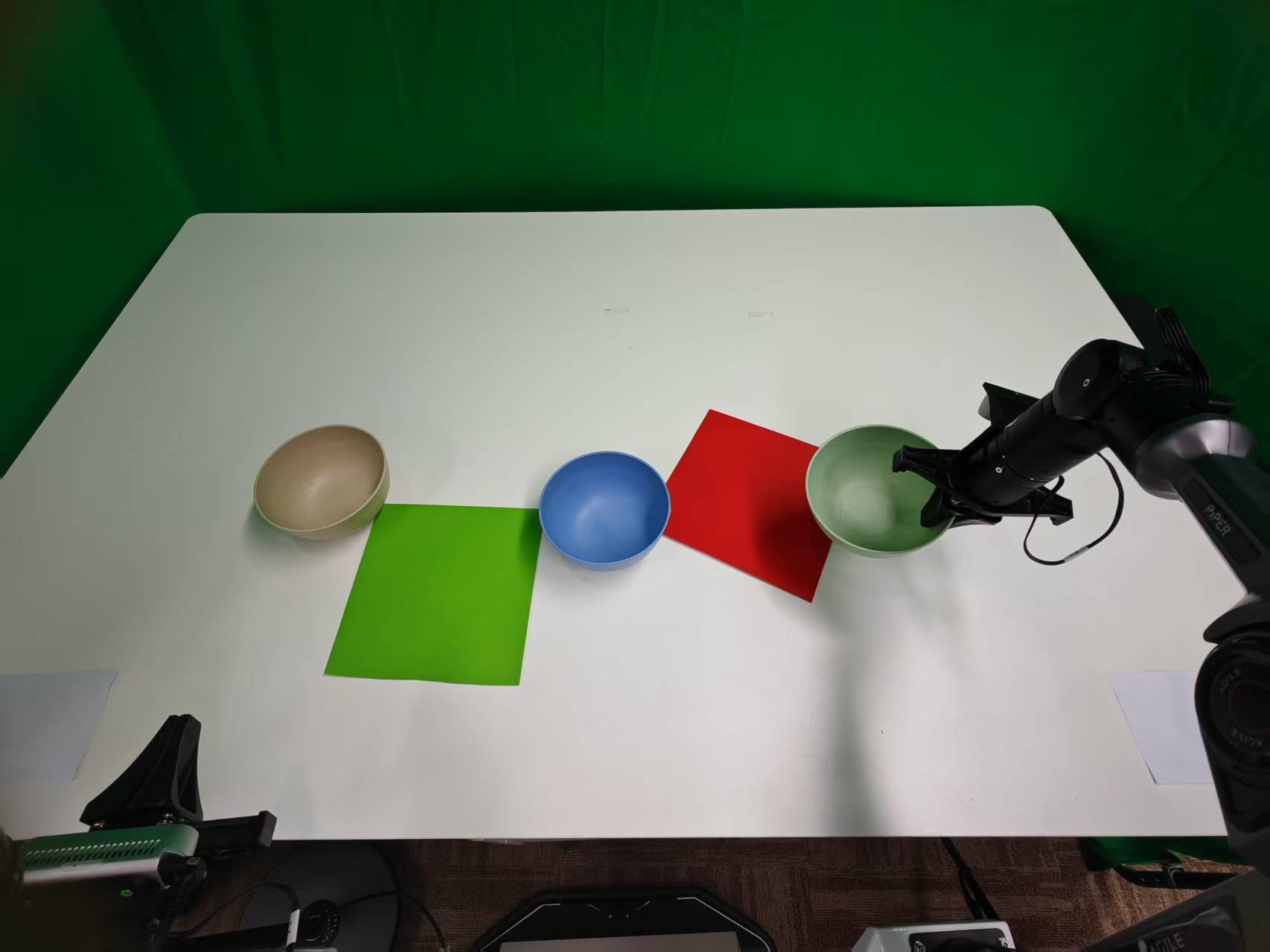}
        & \includegraphics[width=0.30\columnwidth]{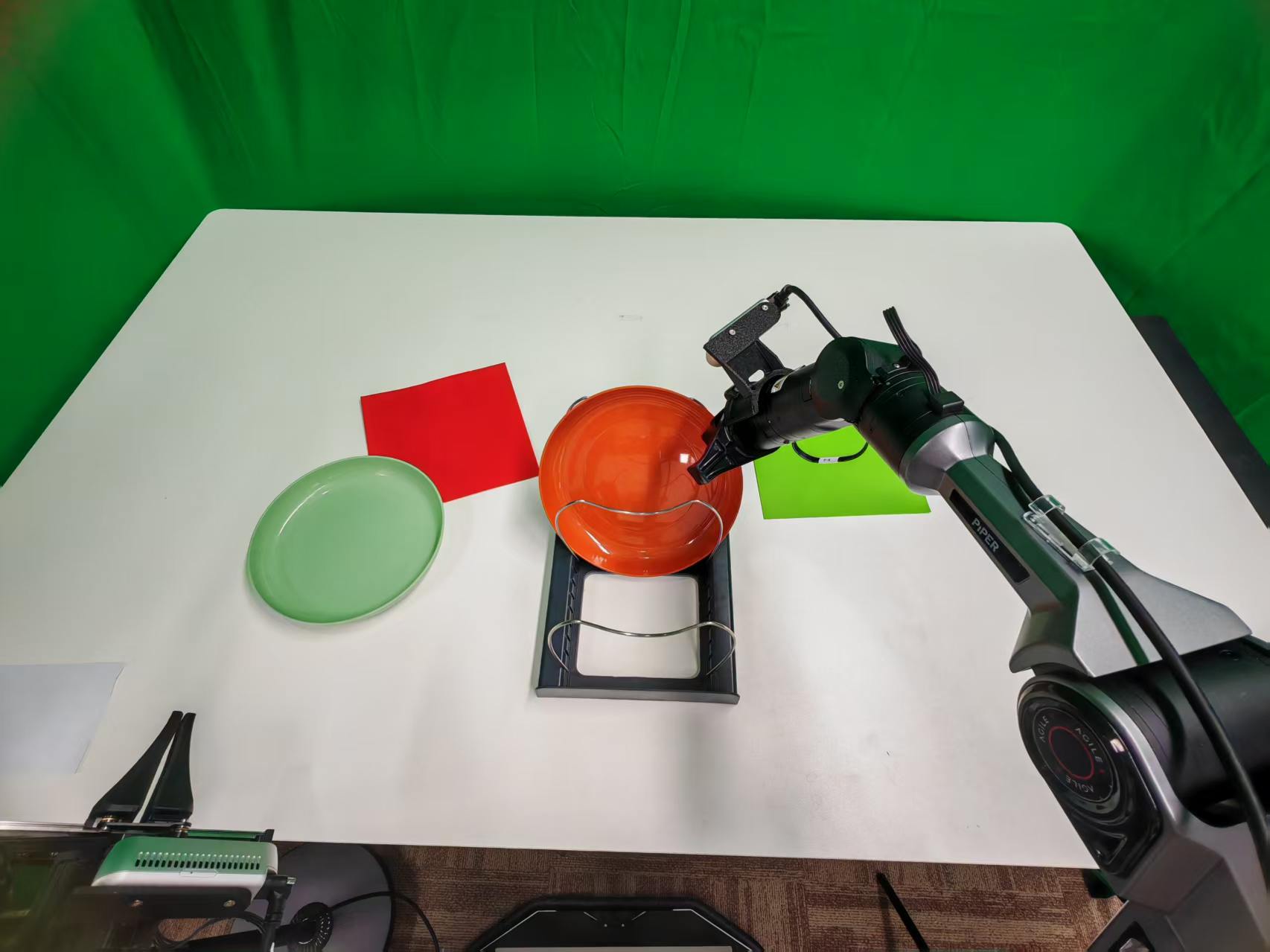}
        & \includegraphics[width=0.30\columnwidth]{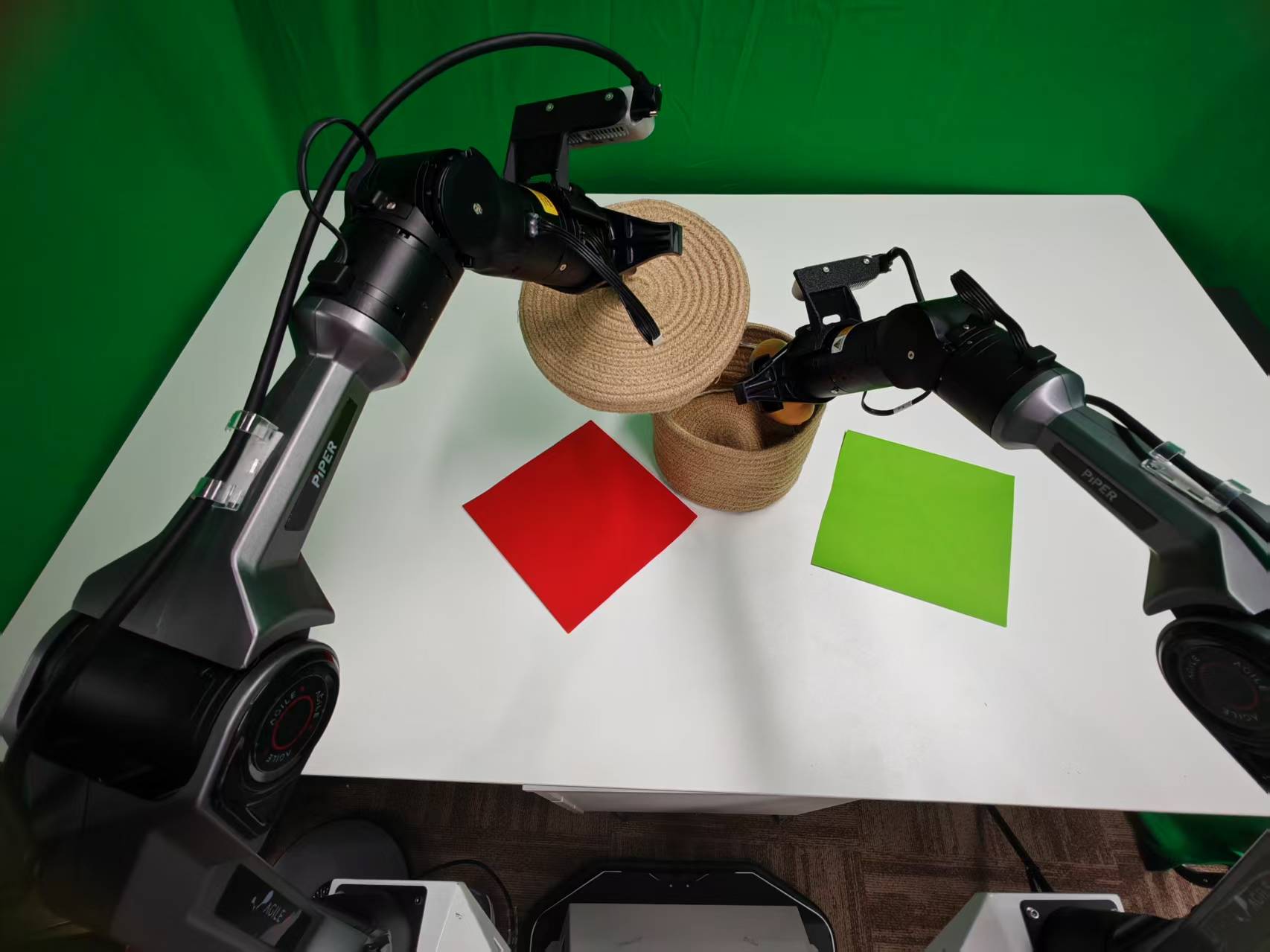}
    \end{tabular}%

    \caption{
    Real-world task-by-condition evaluation matrix.
    Columns show three bimanual manipulation tasks, while rows show
    clean, lighting-perturbed, and background-perturbed settings.
    }
    \label{fig:real_world_matrix}
\end{figure}

We evaluate DC-WAM on the Agilex Piper bimanual platform shown in Fig.~\ref{fig:real_world_matrix} using three long-horizon tasks: T1, stacking three bowls; T2, stacking plates on a shelf; and T3, opening a basket, placing a potato inside, and closing it. We collect 100 successful demonstrations per task. Tracker-derived dynamic maps are generated offline from training videos and used only as training supervision; execution requires no external tracker, segmentation model, or ground-truth dynamic map.

For each method, we conduct 100 trials per task–condition pair under clean, lighting, and background conditions. Lighting is varied using an external spotlight, while colored paper patches serve as localized background distractors, altering appearance without substantially changing the underlying dynamics.

\subsection{Main Results}

\paragraph{Main Results on LIBERO and LIBERO-Plus.}

Table~\ref{tab:main_ood} reports ID performance on LIBERO and OOD robustness on LIBERO-Plus. DC-WAM reaches 98.1\% on LIBERO, improving FastWAM-AC by 1.4 points and FastWAM by 0.5 points. On LIBERO-Plus, it achieves 60.9\%, outperforming the two baselines by 7.1 and 9.4 points, respectively. DC-WAM also has the smallest ID--OOD drop (37.2 versus 42.9 and 46.1 points) and improves six of seven perturbation dimensions, with the largest gains under language, background, and lighting shifts.





\begin{table}[t]
\centering
\small
\setlength{\tabcolsep}{4.0pt}
\begin{tabular}{@{}c|l|ccc@{}}
\toprule
\multicolumn{2}{c|}{Evaluation}
& FastWAM
& {FastWAM-AC}
& \textbf{DC-WAM} \\
\midrule
\textit{LIBERO}
& Avg.
& 97.6 & 96.7 & \textbf{98.1} \\
\midrule
\multirow{8}{*}{%
{%
        \shortstack{\textit{LIBERO-Plus}\\\textit{(OOD)}}}}
& Camera
& 16.2 & \textbf{24.0} & 23.9 \\
& Robot
& 44.9 & 43.9 & \textbf{51.7} \\
& Lang. 
& 70.9 & 70.7 & \textbf{83.4} \\
& Light
& 80.8 & 81.4 & \textbf{91.7} \\
& Bg.
& 54.1 & 50.5 & \textbf{61.3} \\
& Noise
& 40.5 & 50.8 & \textbf{54.2} \\
& Layout
& 63.4 & 63.2 & \textbf{69.8} \\
& Avg.
& 51.5 & 53.8 & \textbf{60.9} \\
\midrule
\multicolumn{2}{l|}{ID--OOD drop $\downarrow$}
& 46.1 & 42.9 & \textbf{37.2} \\
\bottomrule
\end{tabular}
\caption{
Success rate (\%) on LIBERO (ID) and LIBERO-Plus (OOD), with all methods
trained only on clean LIBERO. ID--OOD drop denotes the difference between
average success rates; lower is better.
}
\label{tab:main_ood}
\end{table}

\paragraph{Results on Real-world Experiments.}

Table~\ref{tab:real_world_results} reports real-world performance for policies trained only on clean demonstrations. DC-WAM consistently improves success across the three tasks, with larger gains under lighting shifts and background perturbations. It also exhibits smaller clean-to-OOD performance drops, demonstrating improved real-world OOD robustness.

\begin{table}[t]
\centering
\small
\setlength{\tabcolsep}{2pt}
\begin{tabular}{@{}llccc@{}}
\toprule
Task & Setting & FastWAM-AC & DC-WAM & $\Delta$ \\
\midrule
\multirow{3}{*}{Stack-Bowl}
 & Clean      & 76 & 84 & 8 \\
 & Light      & 34 ($\downarrow 42$) & 48 ($\downarrow 38$) & 14 \\
 & Background & 52 ($\downarrow 24$) & 61 ($\downarrow 23$) & 9 \\
\midrule
\multirow{3}{*}{Pile-Plates}
 & Clean      & 80 & 89 & 9 \\
 & Light      & 41 ($\downarrow 39$) & 71 ($\downarrow 18$) & 30 \\
 & Background & 46 ($\downarrow 34$) & 69 ($\downarrow 20$) & 23 \\
\midrule
\multirow{3}{*}{Collect-Potato}
 & Clean      & 62 & 70 & 8 \\
 & Light      & 53 ($\downarrow 9$) & 66 ($\downarrow 4$) & 13 \\
 & Background & 30 ($\downarrow 32$) & 45 ($\downarrow 25$) & 15 \\
\bottomrule
\end{tabular}

\caption{Real-world success rate (\%).
All methods are trained only on clean demonstrations and evaluated
without adaptation under clean and unseen appearance-level
perturbations. Values in parentheses indicate the absolute
success-rate drop relative to the corresponding clean condition.}
\label{tab:real_world_results}
\end{table}

\subsection{Ablation Studies}
\label{sec:ablations}

\paragraph{Dense and sparse dynamic supervision.}

We ablate dense and sparse dynamic supervision under the same DynaRoute-enabled setting. Dense corresponds to temporal-difference supervision over the full VAE latent grid, while Sparse corresponds to TrackFM reweighting around tracker-derived interaction regions. As shown in Table~\ref{tab:supervision_ablation}, each signal improves over FastWAM-AC, and their combination in DC-WAM performs best, indicating that global temporal changes and localized interaction dynamics are complementary.

\begin{table}[t]
\centering
\small
\setlength{\tabcolsep}{4pt}
\begin{tabular}{lcc}
\toprule
Variant & LIBERO & LIBERO-Plus \\
\midrule
FastWAM-AC & 96.7 & 53.8 \\
+ Dense w/ Route & 97.1 & 56.6 \\
+ Sparse w/ Route & 97.4 & 58.7 \\
DC-WAM & \textbf{98.1} & \textbf{60.9} \\
\bottomrule
\end{tabular}
\caption{
Ablation of dense and sparse dynamic supervision with DynaRoute enabled.
Dense denotes temporal-difference supervision, and sparse denotes TrackFM supervision.
FastWAM-AC removes both dynamic supervision and DynaRoute.
}
\label{tab:supervision_ablation}
\end{table}

\begin{table}[t]
\centering
\small
\setlength{\tabcolsep}{4pt}
\begin{tabular}{lcc}
\toprule
Routing variant & LIBERO & LIBERO-Plus \\
\midrule
No routing & 97.7 & 59.1 \\
Shuffled relevance & 96.8 & 57.8 \\
Action-query routing & 95.8 & 49.2 \\
DynaRoute & \textbf{98.1} & \textbf{60.9} \\
\midrule
DynaRoute w/ stop-grad & 97.0 & 57.0 \\
\bottomrule
\end{tabular}
\caption{
Ablation of dynamics-aware attention bias.
}
\label{tab:routing_ablation}
\end{table}

\paragraph{Dynamics-aware routing and gradient path.}

Table~\ref{tab:routing_ablation} compares routing strategies under the same dynamic supervision. DynaRoute applies a key-side bias to visual-token attention, while the variants remove routing, shuffle relevance, or bias action-query attention. Its gains over no routing and shuffled relevance demonstrate the importance of spatially aligned dynamic cues, whereas the degradation of action-query routing suggests that these cues should guide the visual branch rather than action tokens directly.

To further examine the visual-to-action gradient path, we use a stop-gradient variant that blocks gradients from visual objectives to the action branch while preserving the forward computation. The resulting drops of 1.1 points on LIBERO and 3.9 points on LIBERO-Plus indicate that video supervision contributes to action learning through the action-conditioned visual computation path.

\subsection{Analysis}

\paragraph{Does higher PSNR imply better control?}

Figure~\ref{fig:sr_psnr_analysis} shows that future-frame PSNR does not correlate monotonically with policy success across supervision variants. Dense w/ Route yields the lowest PSNR but still outperforms FastWAM-AC, while DC-WAM achieves the highest success despite lower PSNR than several variants. This mismatch indicates that standard video-quality metrics do not fully capture the control utility of future prediction in WAMs. Focusing supervision and attention on interaction-induced dynamics can better support policy learning, even at the cost of appearance fidelity.

\paragraph{How does DC-WAM reshape visual attention?}

Figure~\ref{fig:case044-clean-noise-3x4}  compares DC-WAM with the matched FastWAM-AC baseline under paired clean and increasingly corrupted observations. As corruption intensifies, FastWAM-AC becomes diffuse and drifts toward irrelevant background regions, whereas DC-WAM consistently attends to dynamic-relevant objects and interaction regions. This suggests that dynamic-centric supervision and relevance routing produce robust, change-centric representations that preserve action-relevant cues under visual shifts.


\section{Conclusion}

We introduced DC-WAM, a dynamic-centric World-Action Model that improves the control utility of future prediction by shifting the RGB video branch from appearance reconstruction toward interaction-induced dynamics. DC-WAM combines dense and sparse visual objectives with DynaRoute attention bias, requires no additional modality-specific prediction, and supports efficient action-only inference. Experiments on LIBERO, LIBERO-Plus, and real-world bimanual tasks demonstrate improved success and robustness. Ablations confirm the complementarity of dense and sparse supervision, the effectiveness of dynamics-aware routing, and the mismatch between PSNR and control performance. Future work will extend DC-WAM to larger datasets, diverse robot platforms, and WAM architectures beyond MoT.

\bibliography{aaai2027}


\end{document}


\raggedbottom
\begin{center}
    {\Large\bfseries \method: Reproducibility and Supplementary Material}\par
    \vspace{3pt}
    {\normalsize Anonymous Submission}
\end{center}
\vspace{8pt}

\section{Datasets and Evaluation Protocols}

\subsection{Simulation Data}

\paragraph{LIBERO.}
We use the four standard LIBERO suites: Spatial, Object, Goal, and Long.
Table~\ref{tab:libero_data} summarizes the training data and preprocessing
pipeline. All methods use the same demonstrations, action space, observation
processing, and clip construction.

\begin{center}
\small
\begin{tabularx}{\linewidth}{@{}lY lY@{}}
\toprule
\textbf{Item} & \textbf{Configuration} & \textbf{Item} & \textbf{Configuration} \\
\midrule
Suites & Spatial, Object, Goal, Long
& Tasks & $4$ suites $\times$ $10$ tasks \\
Filtered episodes & $1{,}712$ ($\approx 43$ per task)
& Data layout & LeRobot v2.1 repositories \\
Views & Third-person + wrist
& Input size & $224\times224$ per view, concatenated to $224\times448$ \\
Action / state & 7-D action; 8-D proprioception
& Clip length & 33 observation steps \\
Video / action horizon & 9 video frames / 32 action steps
& Subsampling ratio & \texttt{action\_video\_freq\_ratio=4} \\
Language & Cached UMT5 embeddings
& Context length & 128 tokens \\
Preprocessing & ToTensor + resize
& Training split & Clean demonstrations only \\
\bottomrule
\end{tabularx}
\captionof{table}{LIBERO training data and preprocessing configuration.}
\label{tab:libero_data}
\end{center}


\paragraph{LIBERO-Plus.}
LIBERO-Plus is used only for zero-shot OOD evaluation. It perturbs the
original LIBERO tasks along seven dimensions, each with difficulty levels
L1--L5. Table~\ref{tab:liberoplus_protocol} reports the size of the full
evaluation protocol used in our implementation.

\begin{center}
\small
\begin{tabular}{@{}lr@{\qquad}lr@{}}
\toprule
\textbf{Perturbation} & \textbf{Tasks}
& \textbf{Perturbation} & \textbf{Tasks} \\
\midrule
Camera viewpoint & 1{,}599 & Lighting & 1{,}142 \\
Robot initialization & 1{,}550 & Background texture & 1{,}076 \\
Language instruction & 1{,}537 & Sensor noise & 1{,}601 \\
Object layout & 1{,}525 & \textbf{Total} & \textbf{10{,}030} \\
\bottomrule
\end{tabular}
\captionof{table}{LIBERO-Plus full-protocol task counts. One trial is executed per
protocol task by default.}
\label{tab:liberoplus_protocol}
\end{center}

\subsection{Real-World Data and Evaluation}

We use an Agilex Piper bimanual platform and collect 100 successful
training demonstrations for each task. All policies are trained only on
clean demonstrations and evaluated without adaptation under clean and
unseen appearance perturbations.

\begin{center}
\small
\begin{tabularx}{\linewidth}{@{}lYcc@{}}
\toprule
\textbf{Task} & \textbf{Description} & \textbf{Train demos} & \textbf{Eval conditions} \\
\midrule
Stack-Bowl & Stack three bowls at the table center. & 100 & Clean / Light / Bg. \\
Pile-Plates & Pick up plates and place them on the shelf. & 100 & Clean / Light / Bg. \\
Collect-Potato & Open the basket, insert the potato, and close it. & 100 & Clean / Light / Bg. \\
\bottomrule
\end{tabularx}
\captionof{table}{Real-world task and data summary.}
\label{tab:real_world_data}
\end{center}

\begin{center}
\begin{minipage}{\linewidth}
\centering

\textbf{T1}\quad Stack the three bowls at the center of the table.
\par\vspace{3pt}
\noindent\hbox to \linewidth{%
    \includegraphics[width=0.163\linewidth]{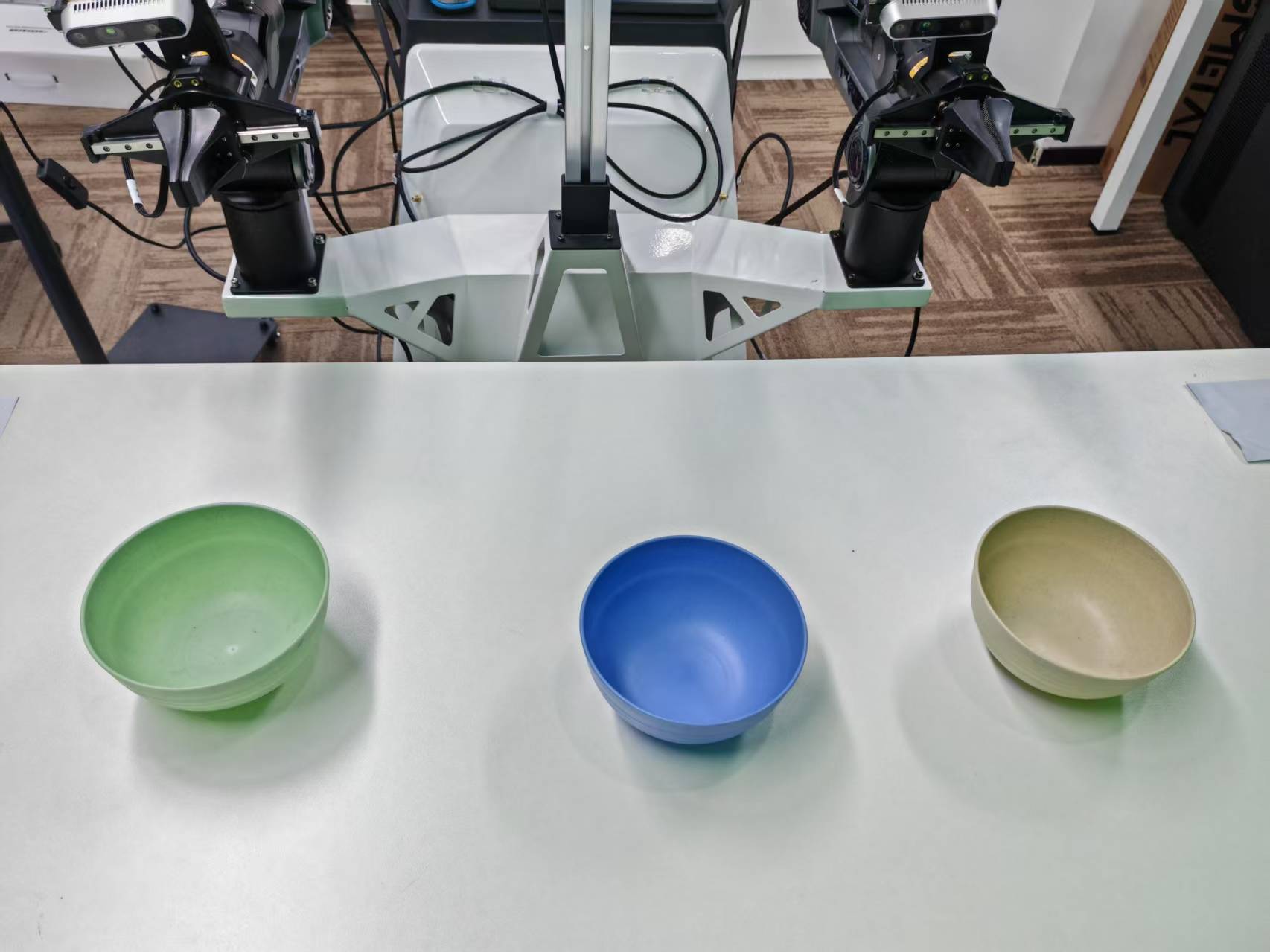}\hfil
    \includegraphics[width=0.163\linewidth]{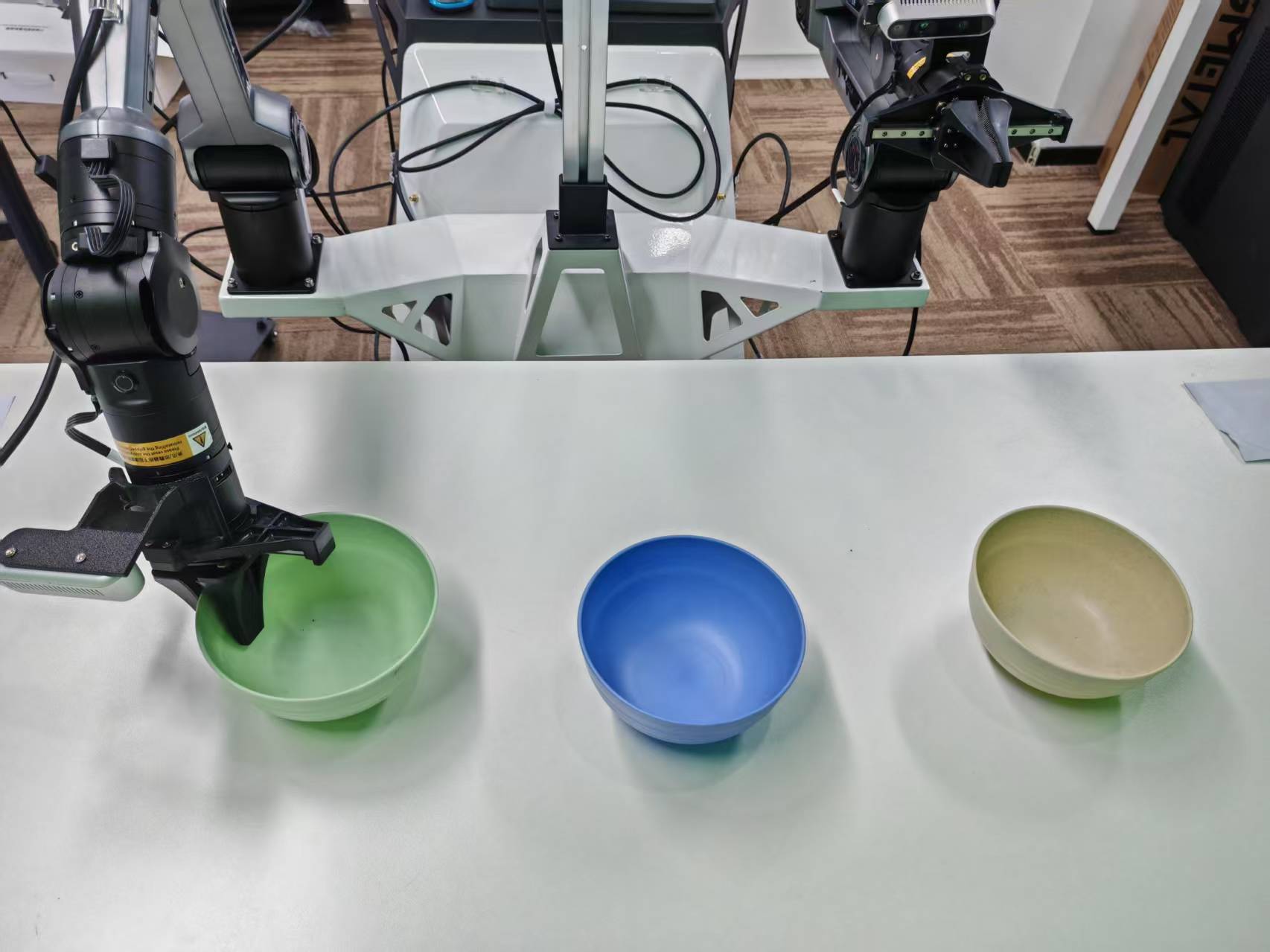}\hfil
    \includegraphics[width=0.163\linewidth]{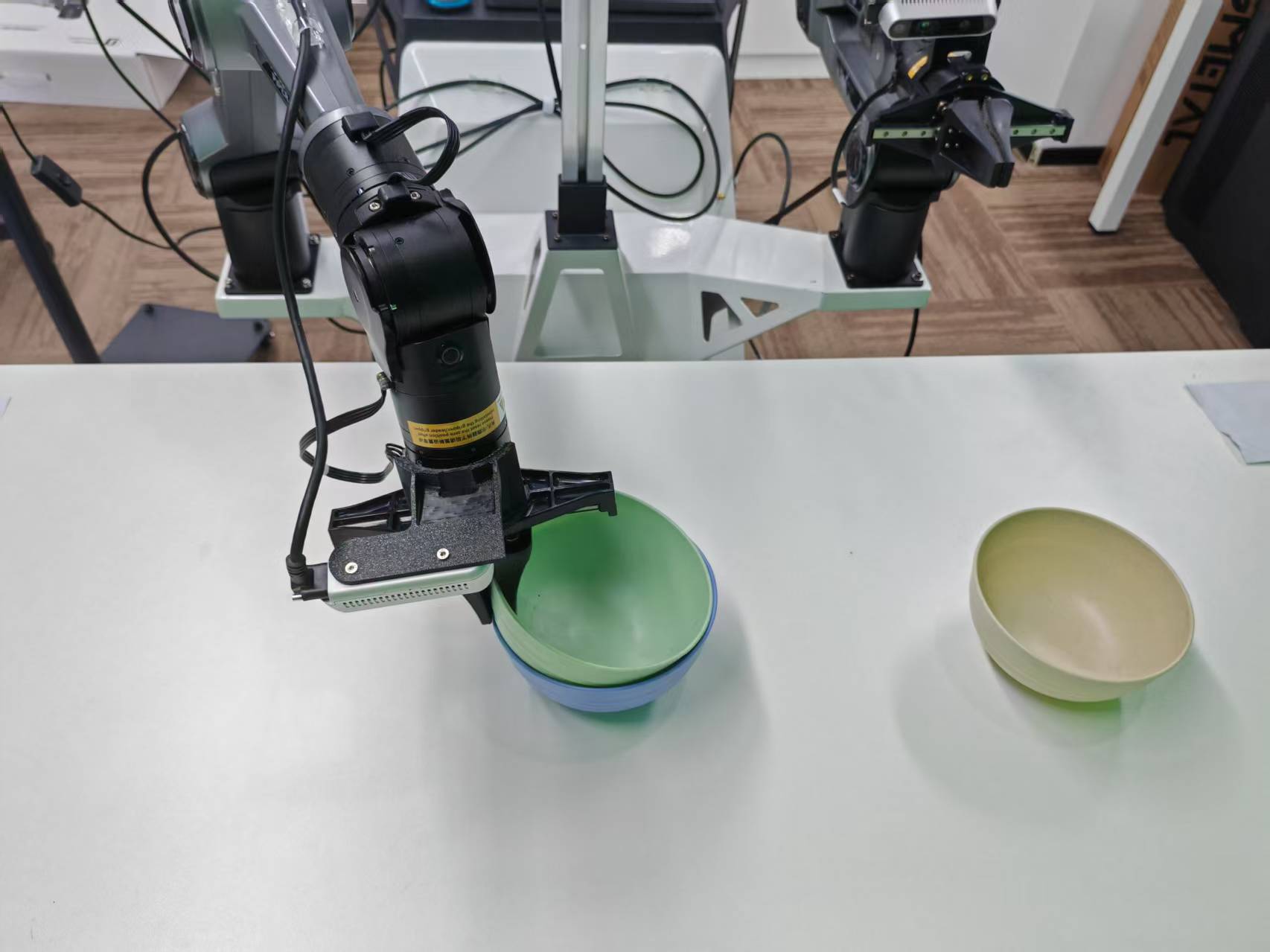}\hfil
    \includegraphics[width=0.163\linewidth]{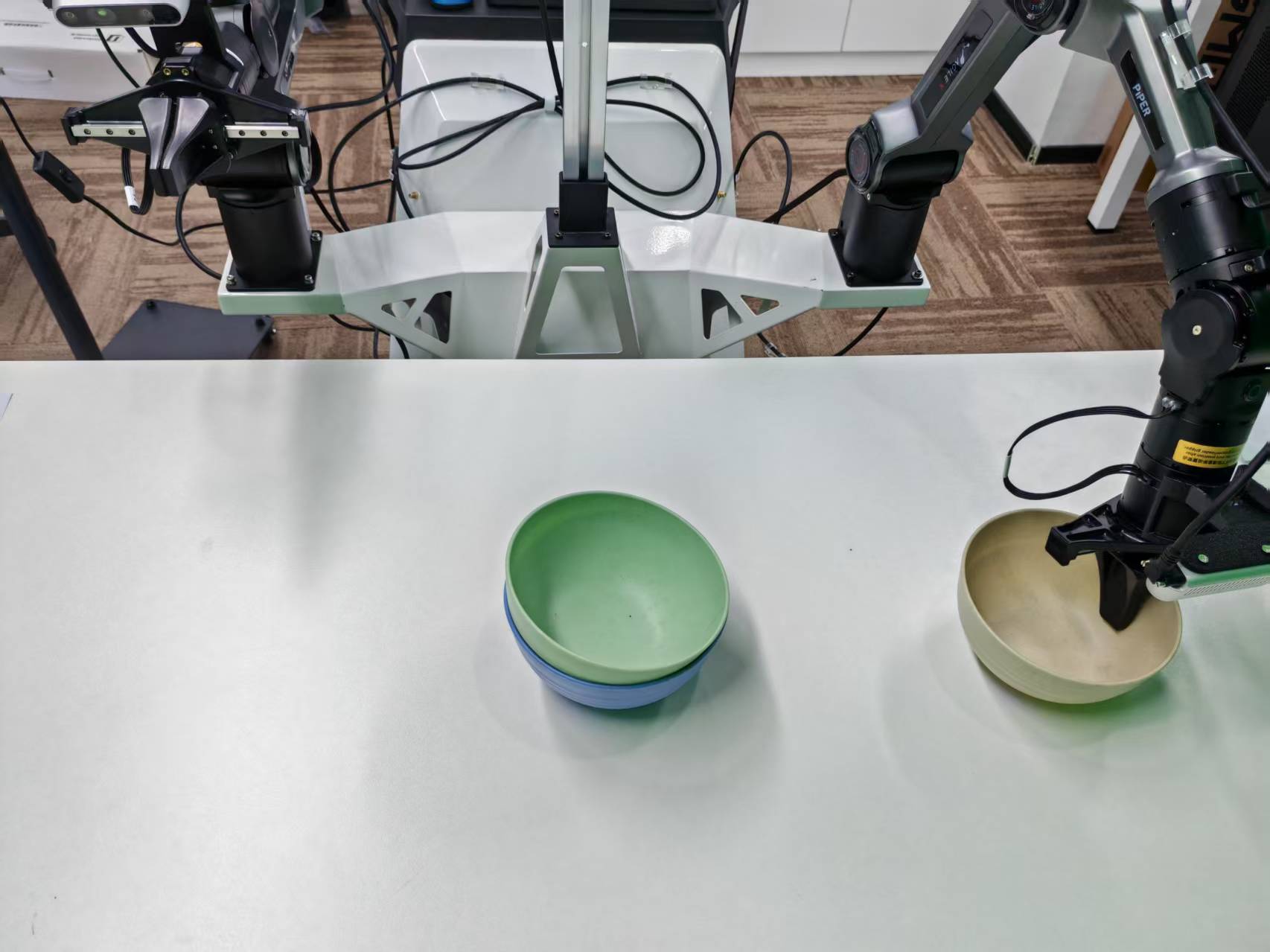}\hfil
    \includegraphics[width=0.163\linewidth]{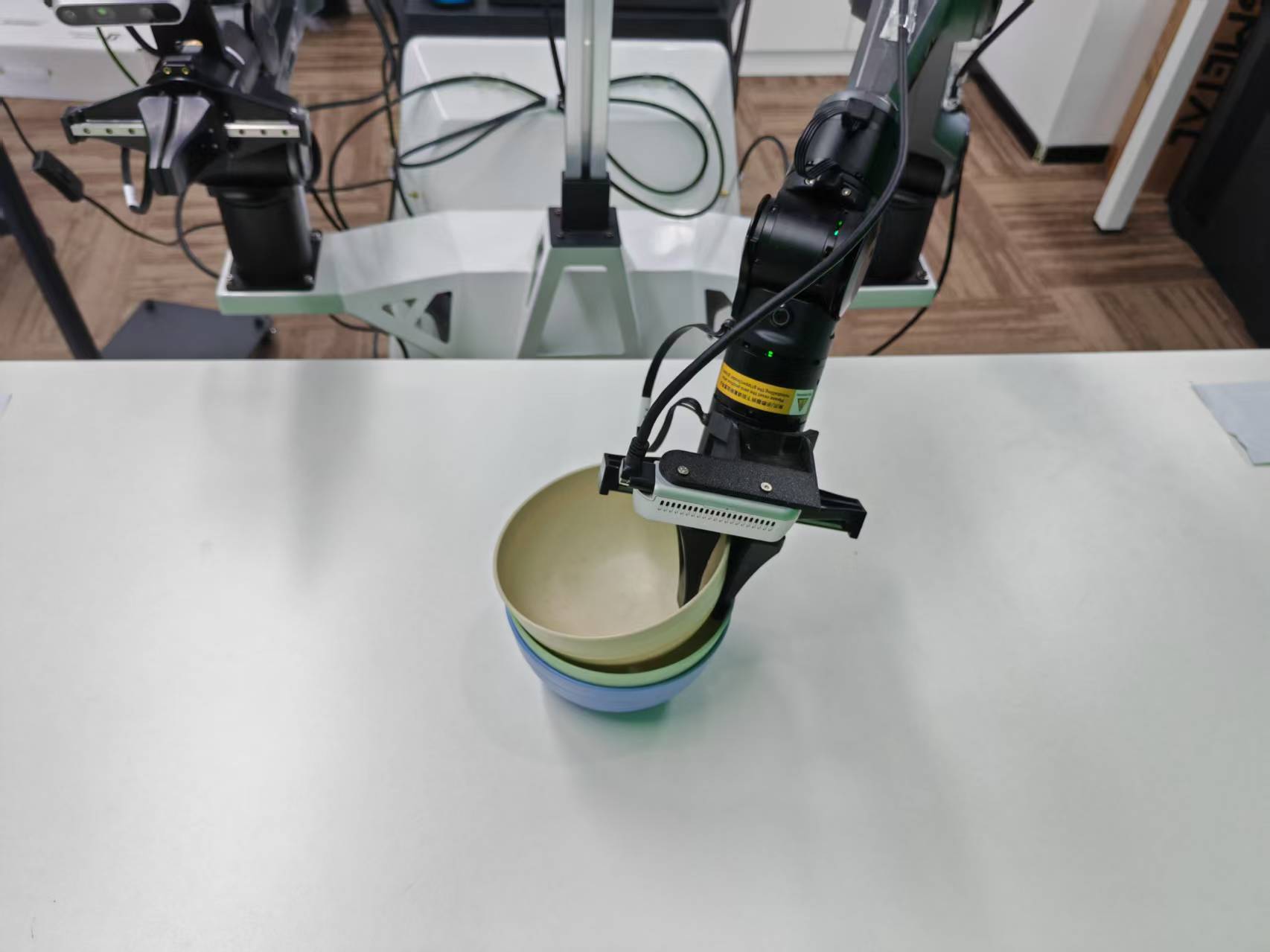}\hfil
    \includegraphics[width=0.163\linewidth]{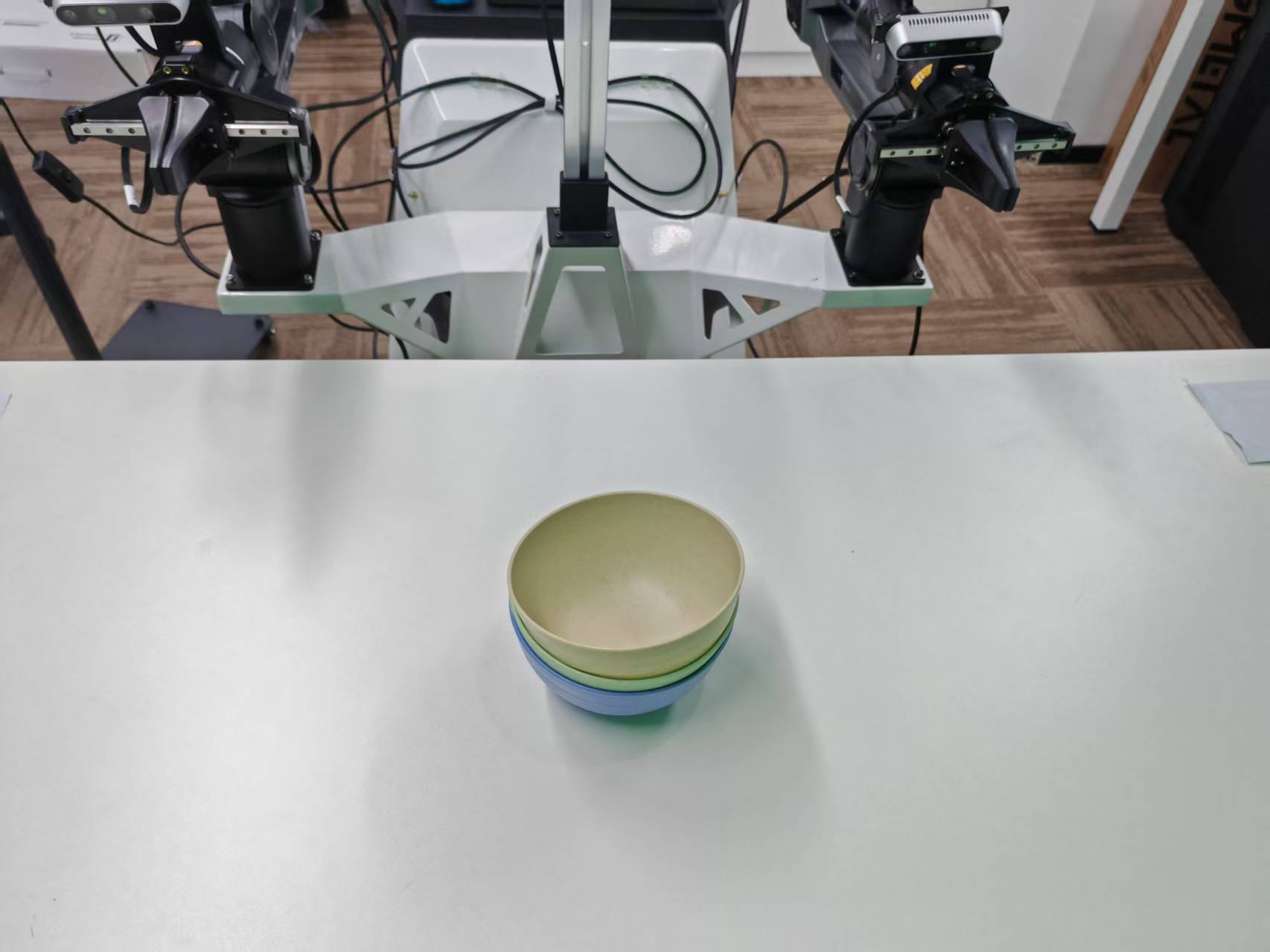}%
}

\par\vspace{6pt}
\textbf{T2}\quad Pick up the plates and place them onto the shelf.
\par\vspace{3pt}
\noindent\hbox to \linewidth{%
    \includegraphics[width=0.163\linewidth]{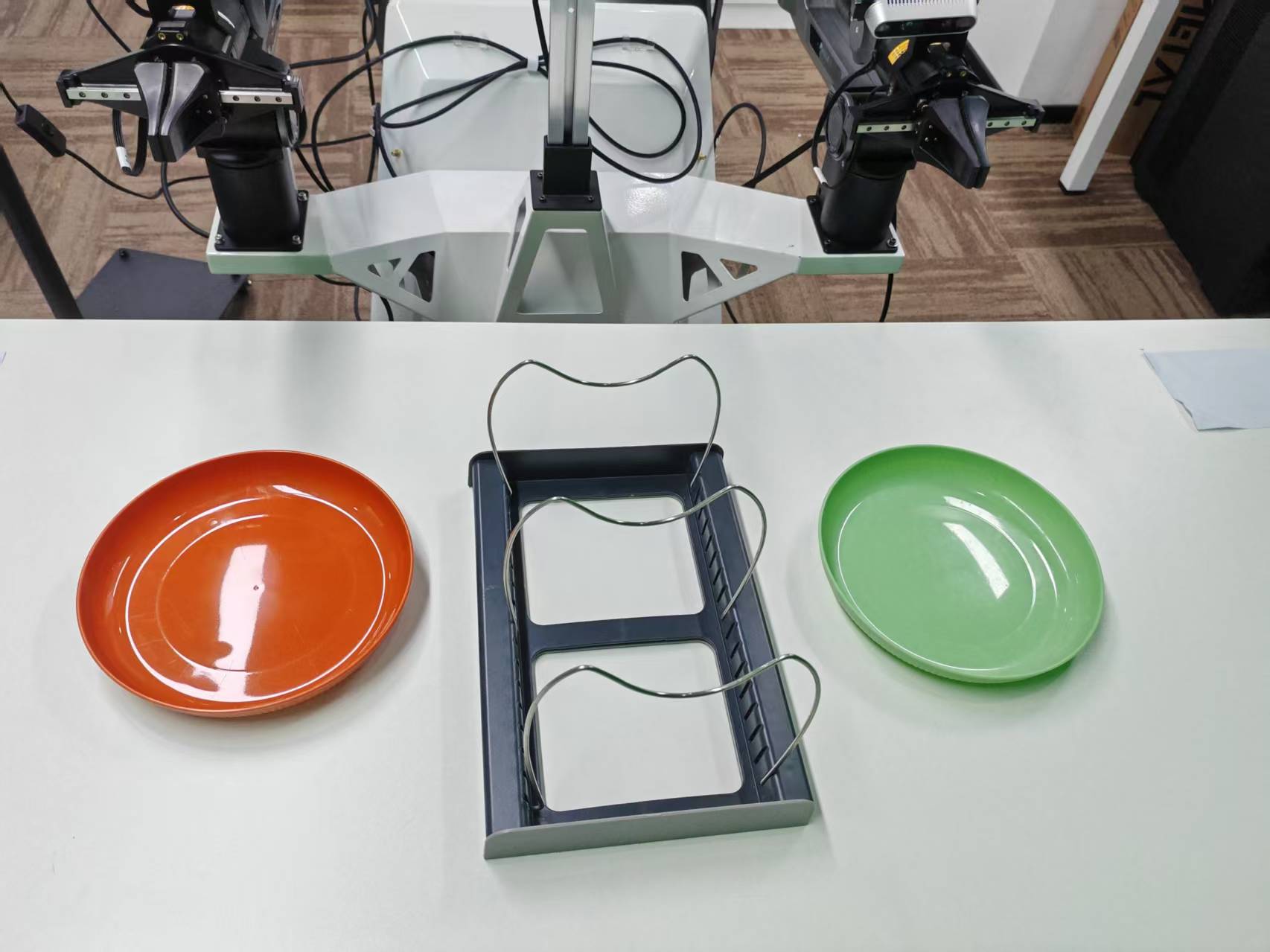}\hfil
    \includegraphics[width=0.163\linewidth]{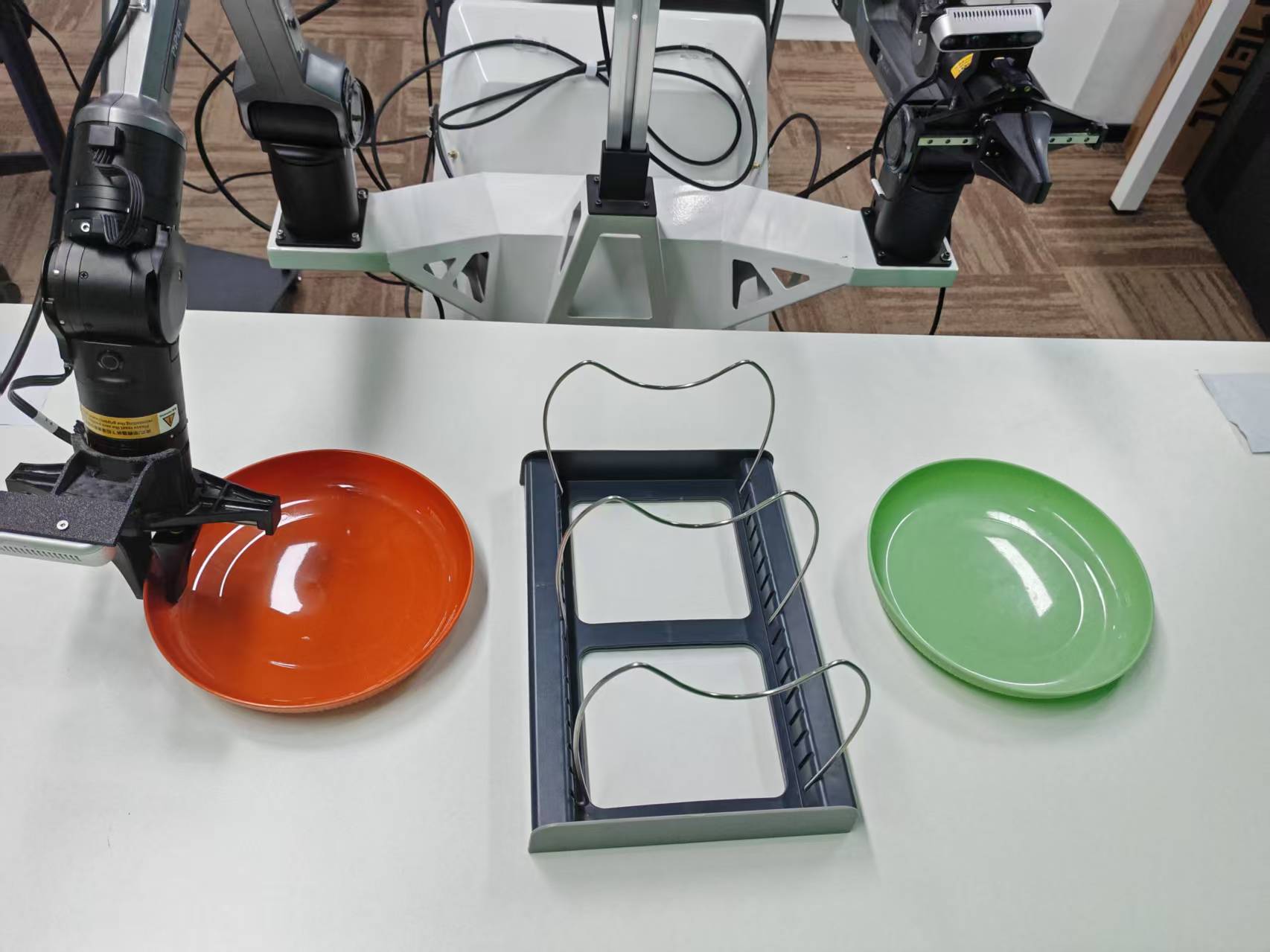}\hfil
    \includegraphics[width=0.163\linewidth]{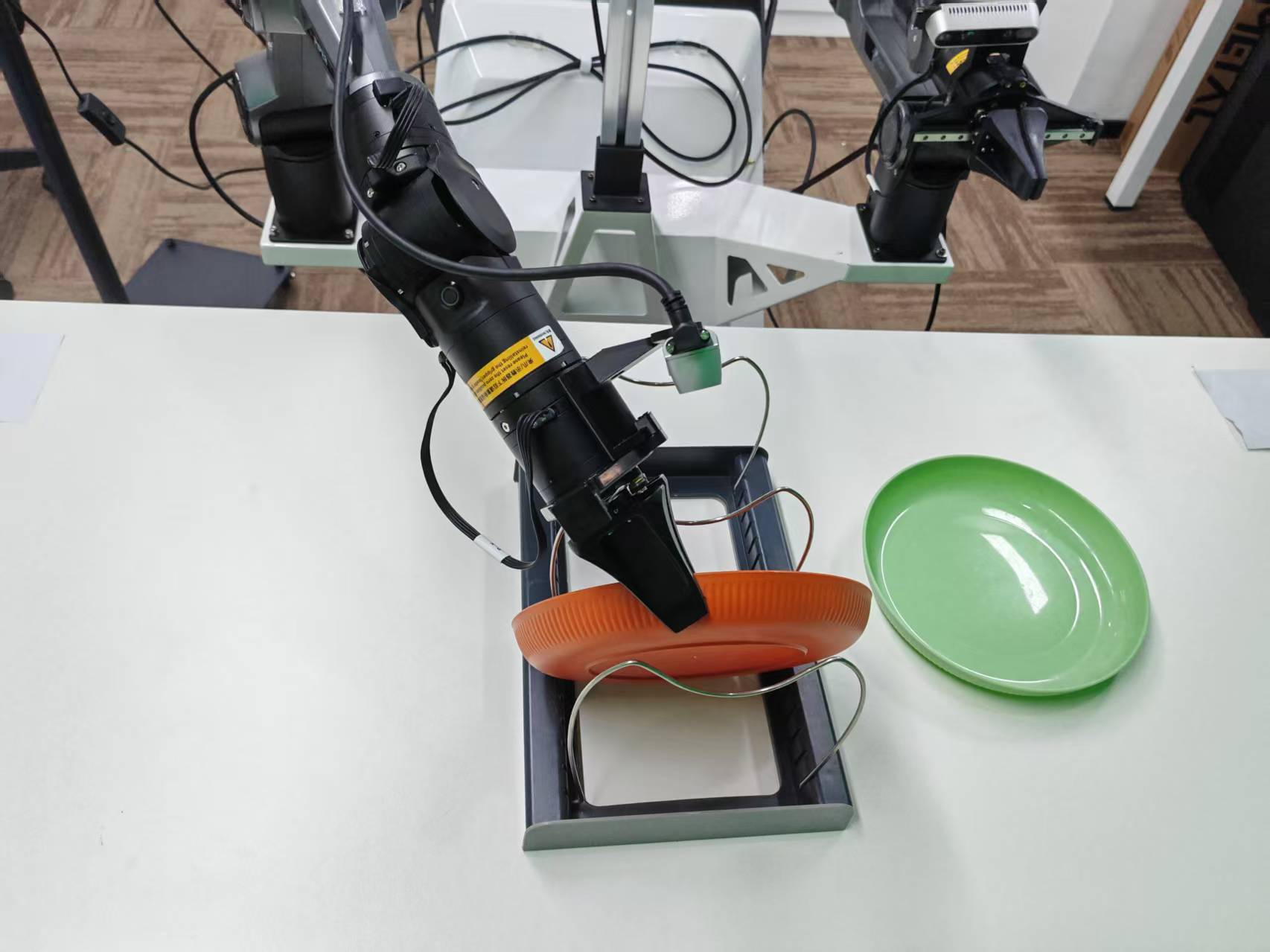}\hfil
    \includegraphics[width=0.163\linewidth]{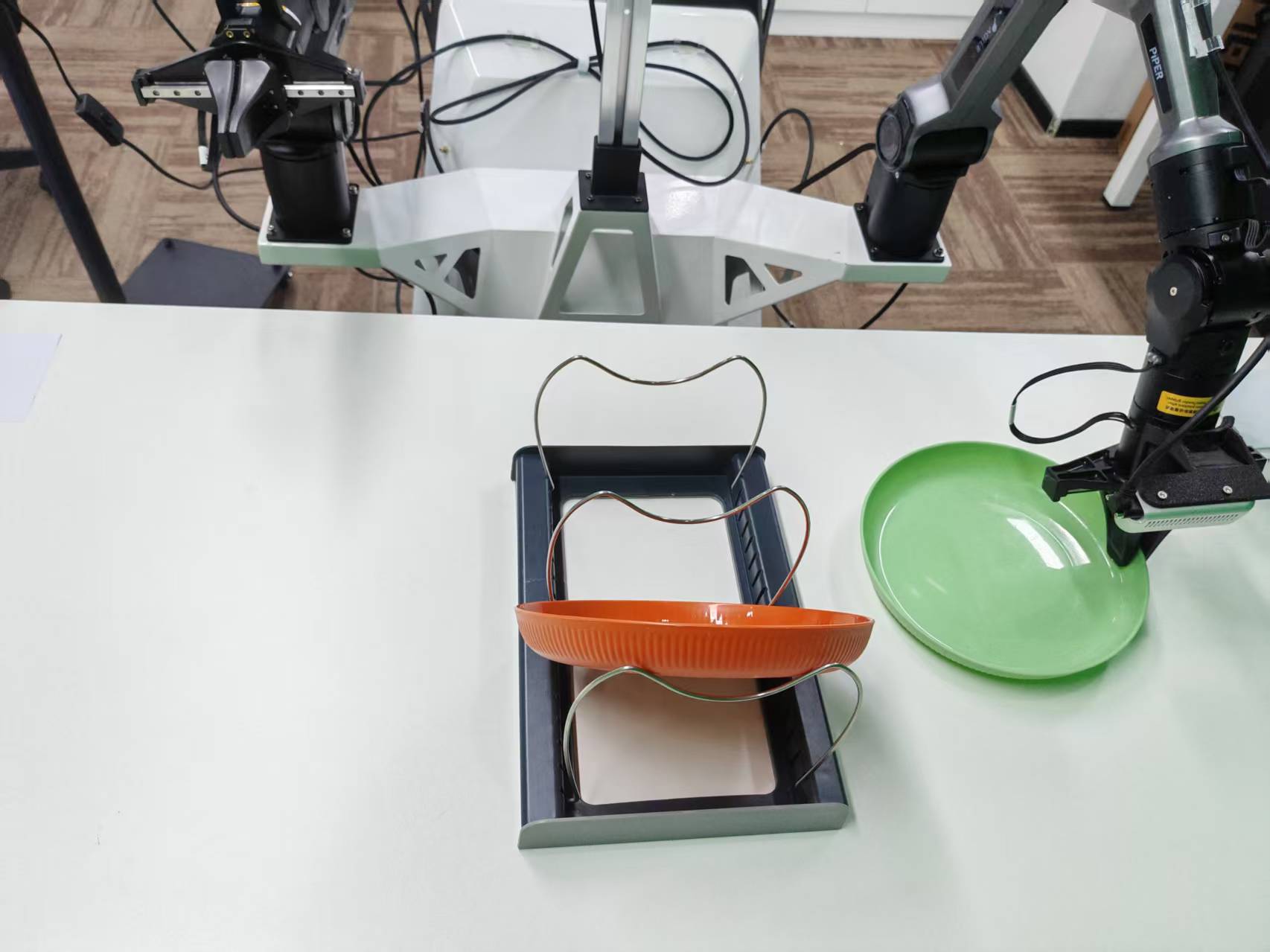}\hfil
    \includegraphics[width=0.163\linewidth]{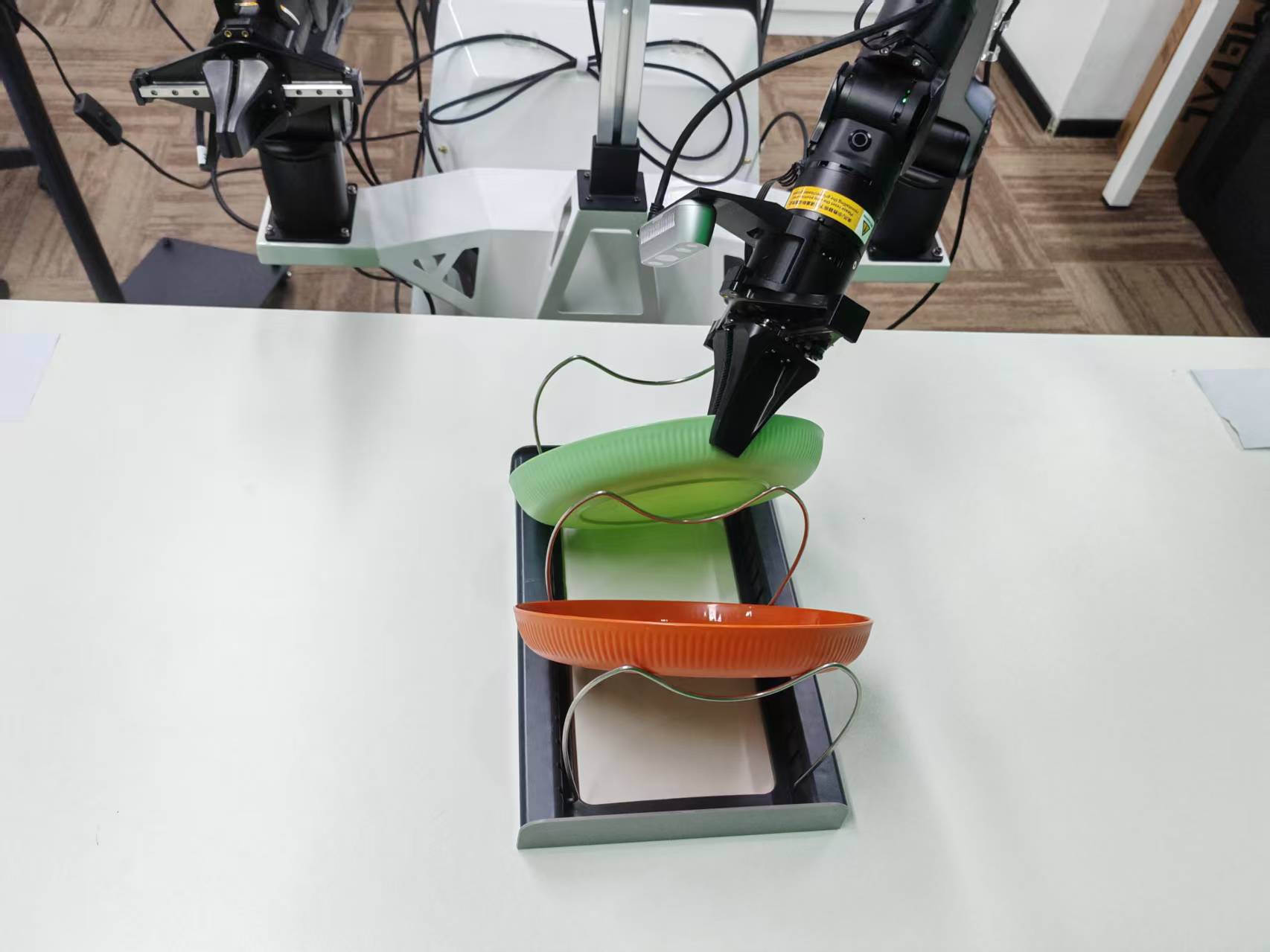}\hfil
    \includegraphics[width=0.163\linewidth]{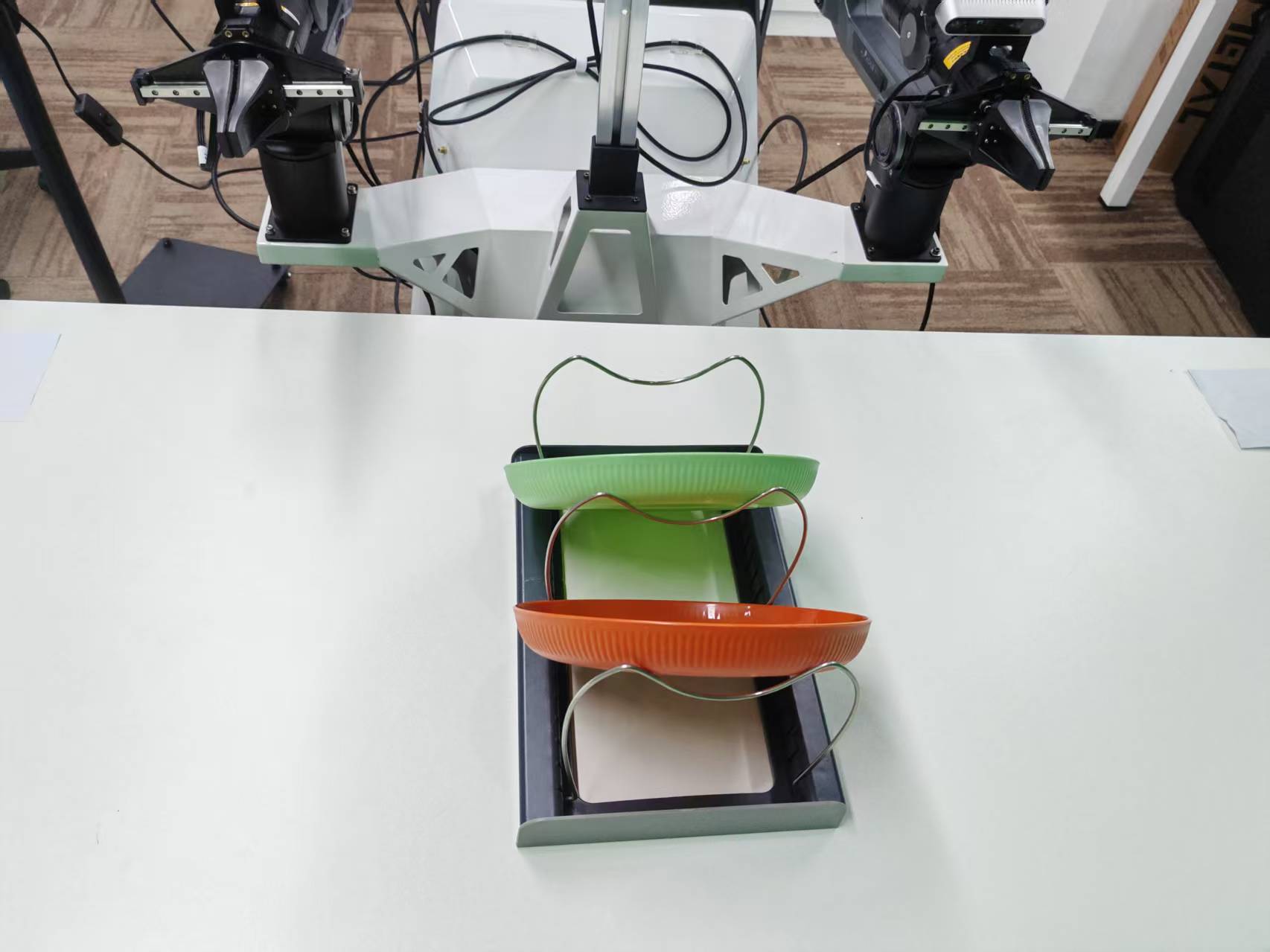}%
}

\par\vspace{6pt}
\textbf{T3}\quad Open the basket, place the potato inside, and close it.
\par\vspace{3pt}
\noindent\hbox to \linewidth{%
    \includegraphics[width=0.163\linewidth]{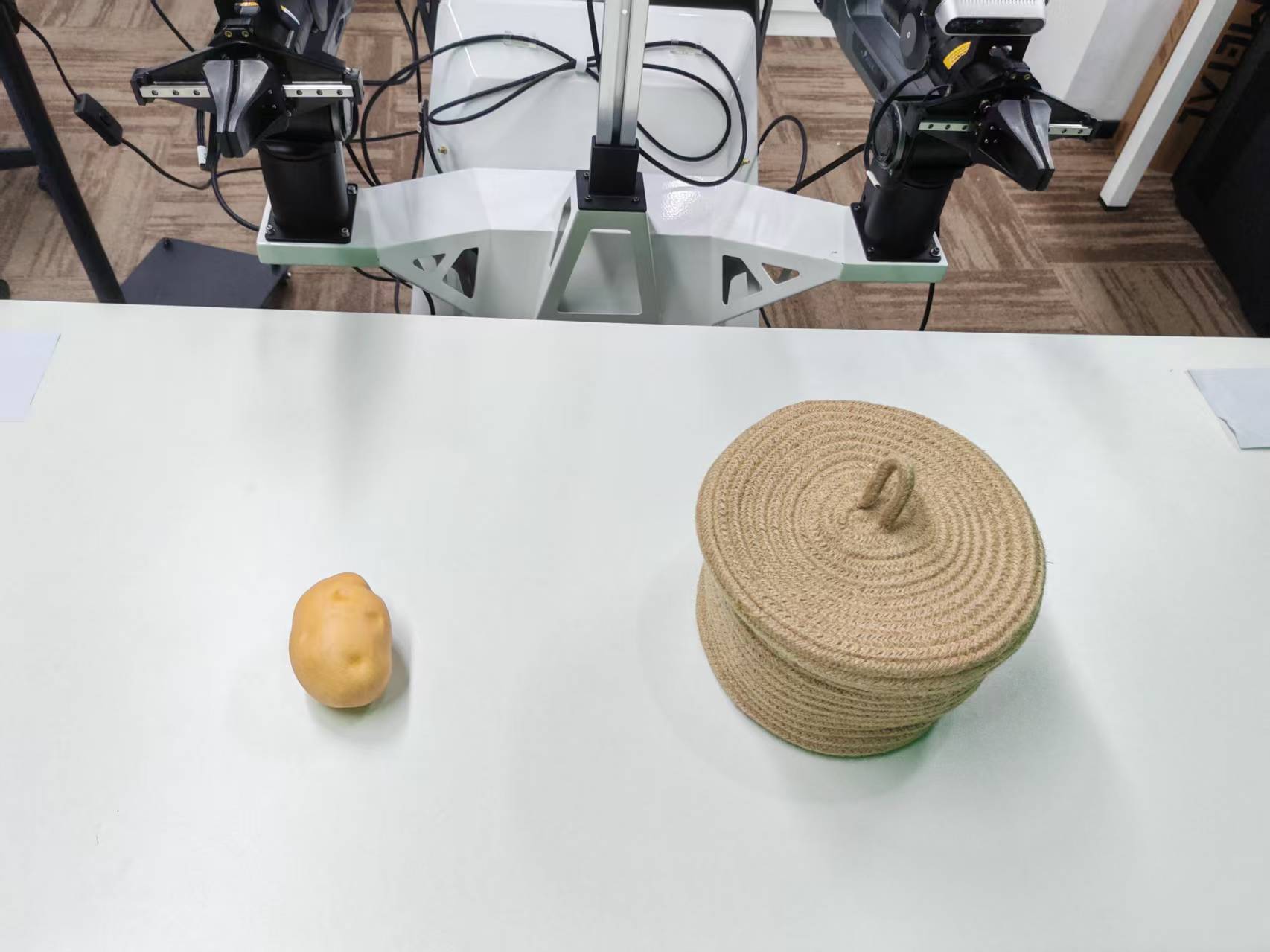}\hfil
    \includegraphics[width=0.163\linewidth]{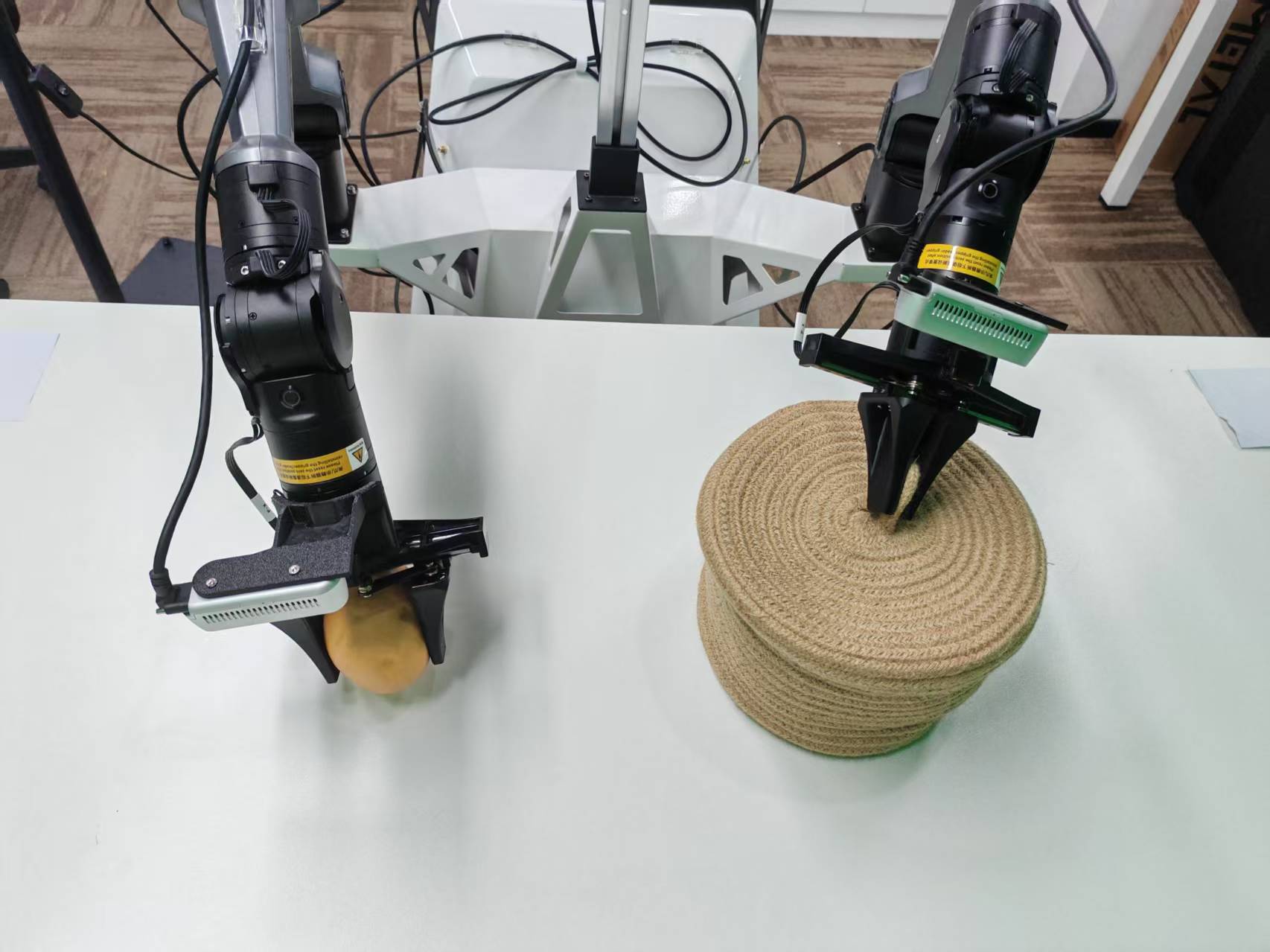}\hfil
    \includegraphics[width=0.163\linewidth]{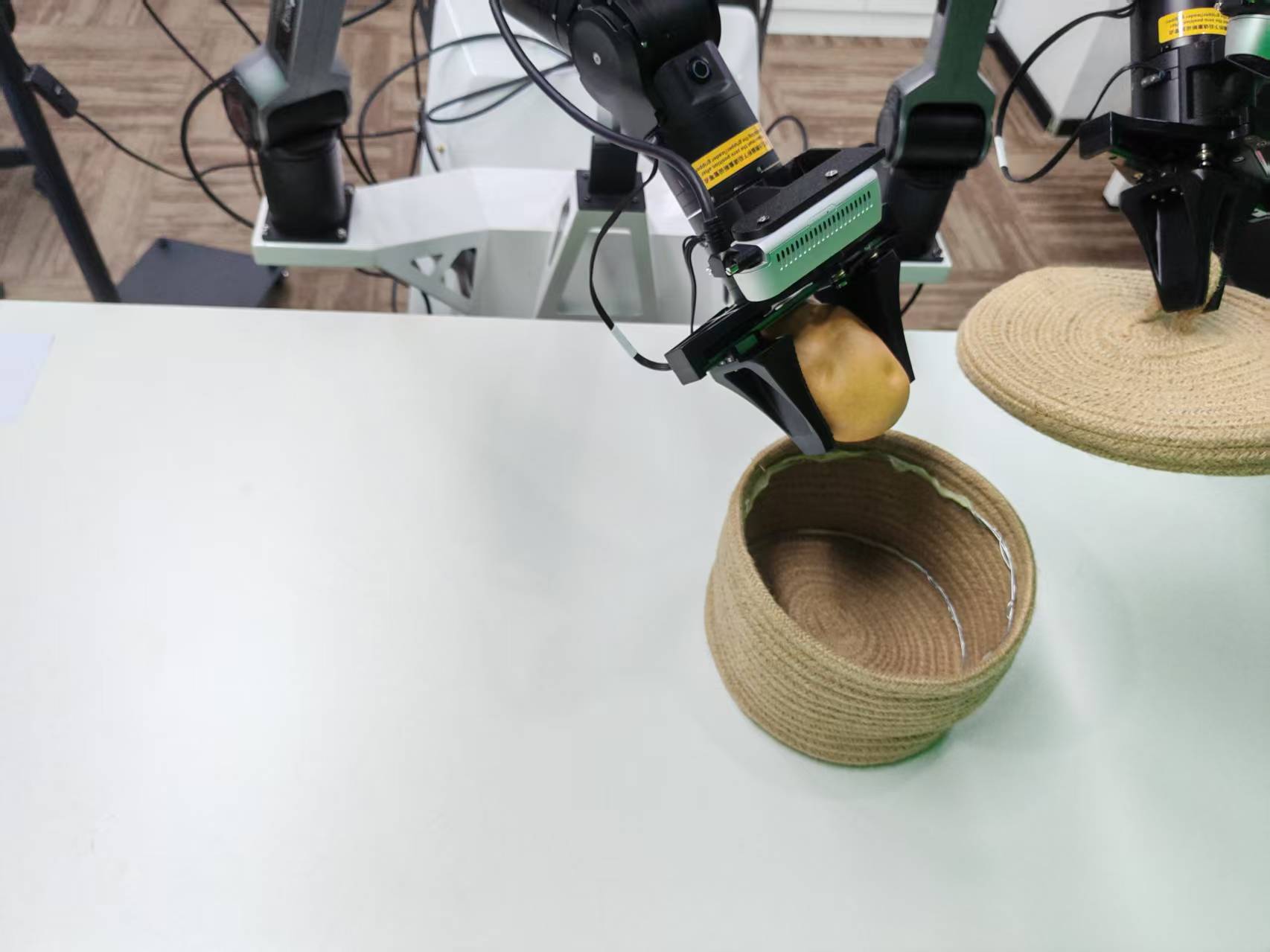}\hfil
    \includegraphics[width=0.163\linewidth]{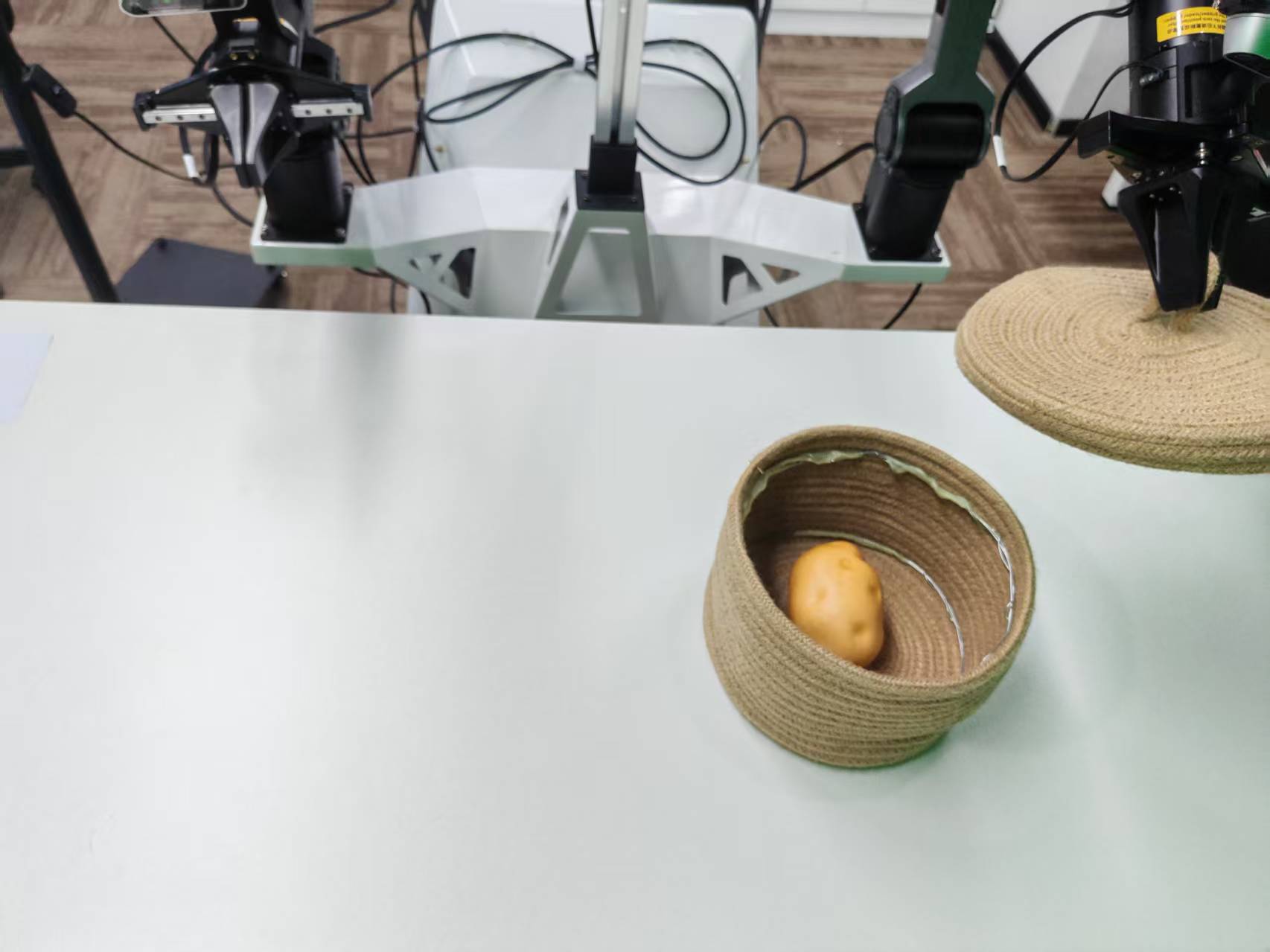}\hfil
    \includegraphics[width=0.163\linewidth]{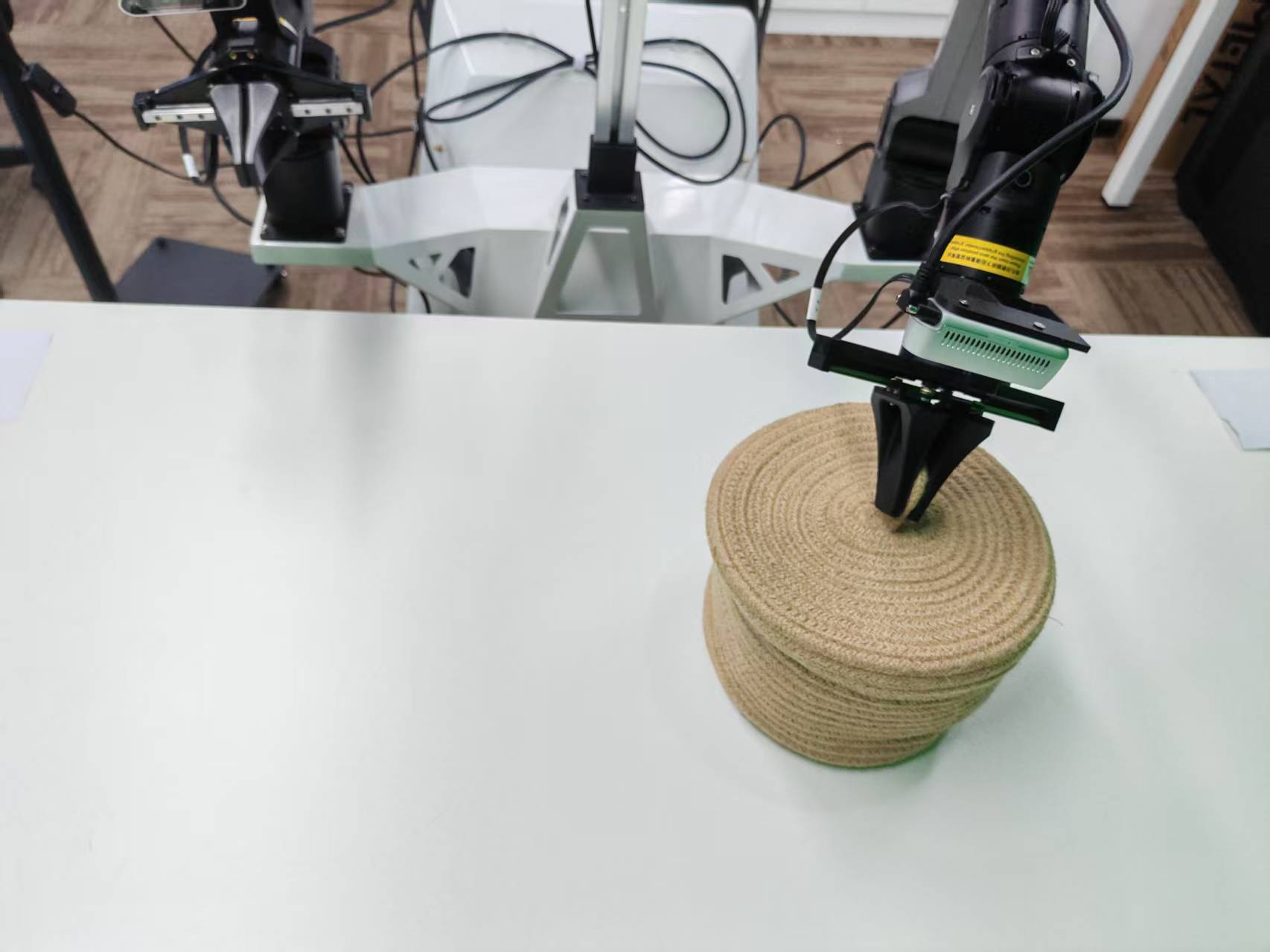}\hfil
    \includegraphics[width=0.163\linewidth]{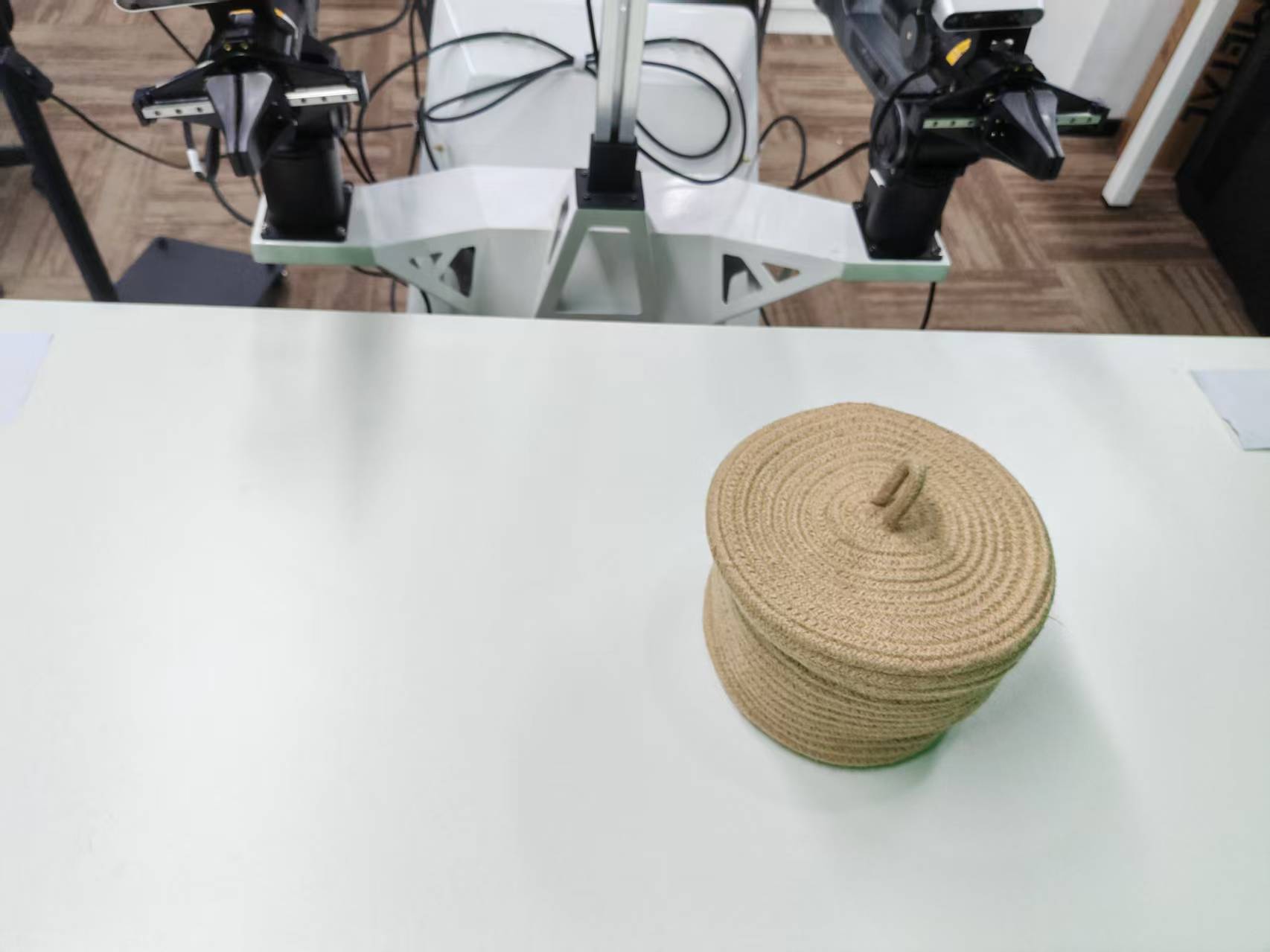}%
}

\captionof{figure}{Representative execution sequences for the three real-world tasks.
Each row shows temporally ordered key frames from one successful rollout.}
\label{fig:real_world_task_sequences}
\end{minipage}
\end{center}

The real-world input contains either three fixed views
($384\times320$ in the three-camera layout) or a front--wrist pair resized and
concatenated to $224\times448$. The bimanual action/state representation is
14-D; the two-camera Stack-Bowl setting uses a 7-D right-arm representation
after dimension selection. Training clips use the same 33-step observation
window and 4:1 action-to-video subsampling ratio as LIBERO.

The lighting condition introduces an external spotlight that changes
illumination and shadows. The background condition adds localized colored
paper distractors without changing the underlying task dynamics. Each
reported task--condition pair is evaluated with 100 independent trials.


\section{Offline Dynamic-Map Construction}

Candidate points are sampled from SAM ViT-B masks using a hybrid grid and
representative-point strategy. CoTracker3 runs offline at the episode level;
no tracker or segmentation model is used during policy execution. The local
tracking window is $T_{\mathrm{local}}=9$, matching the video horizon used for
training.

\begin{center}
\begin{minipage}[t]{0.485\linewidth}
\vspace{0pt}
\centering
\small
\begin{tabularx}{\linewidth}{@{}lY@{}}
\toprule
\textbf{Tracking item} & \textbf{Value} \\
\midrule
Tracker & CoTracker3 offline \\
SAM & ViT-B checkpoint \\
Main-view points & 96 \\
Wrist-view points & 160 \\
Visibility threshold & 0.5 \\
Motion measure & $\lVert\mathbf p_t-\mathbf p_{t-1}\rVert_2$ \\
Motion percentiles & 0.6 / 0.7 / 0.7 \\
Tracking resolution & Native episode RGB \\
\bottomrule
\end{tabularx}
\captionof{table}{Point sampling and tracking configuration.}
\label{tab:tracking_params}
\end{minipage}
\hfill
\begin{minipage}[t]{0.485\linewidth}
\vspace{0pt}
\centering
\small
\begin{tabularx}{\linewidth}{@{}lY@{}}
\toprule
\textbf{Map item} & \textbf{Value} \\
\midrule
Gaussian scale $\lambda$ & 0.25 \\
Kernel width $\sigma_p$ & 1.25 patches \\
VAE grid & $28\times56$ \\
DiT patch size & $[1,2,2]$ \\
Normalization & Per episode and camera view \\
Temporal sampling & Same indices as RGB \\
Latent time steps & $9$ frames $\rightarrow 3$ latent steps \\
Interpolation & Trilinear to DiT token grid \\
\bottomrule
\end{tabularx}
\captionof{table}{Rasterization and normalization configuration.}
\label{tab:rasterization_params}
\end{minipage}
\end{center}

For each visible track, motion is measured as
$d_{t,n}=\lVert\mathbf p_{t,n}-\mathbf p_{t-1,n}\rVert_2$. Points above the
motion threshold are rasterized onto the VAE latent grid with a Gaussian
kernel and normalized separately for each episode and camera view. The map is
sampled using the same temporal indices as the RGB clip and then converted to
the DiT token grid for DynaRoute supervision and attention biasing.

\section{Model and Optimization Details}

\subsection{Architecture}

\begin{center}
\small
\begin{tabularx}{\linewidth}{@{}lY lY@{}}
\toprule
\textbf{Architecture item} & \textbf{Configuration}
& \textbf{DynaRoute item} & \textbf{Configuration} \\
\midrule
Video backbone & Wan2.2-TI2V-5B
& Hidden dimension & 256 \\
Visual--action model & Two-expert MoT
& Fusion attention & 4 heads \\
Visual DiT & $d_v=3072$, 30 layers, 24 heads
& Condition / gate MLP & 2 layers / 2 layers \\
Action DiT & $d_a=1024$, 30 layers, 24 heads
& Time embedding & 256 \\
Current visual input & Clean first-frame latents
& Routing strength $\alpha$ & 0.75 \\
Future visual input & Noised future latents at $\tau$
& Numerical constant $\epsilon$ & $10^{-3}$ \\
Action input & Noised at shared $\tau$
& Route loss weights & BCE 1.0, MSE 0.25, Dice 0.25, focal 0.10 \\
Visual / action horizon & 9 frames / 32 actions
& Layer sharing & One bias vector per denoising step \\
Action dimensions & 7 (sim), 14 (real)
& First-frame key bias & Disabled \\
Conditioning & Language + proprioception
& Input detach / bias detach & True / False \\
\bottomrule
\end{tabularx}
\captionof{table}{Backbone architecture and DynaRoute configuration.}
\label{tab:architecture_dynaroute}
\end{center}

\subsection{DynaRoute Architecture}

The DynaRoute network $G_\psi$ projects visual, action, language, and
proprioceptive features into a shared hidden space, augments visual tokens
with spatiotemporal and token-type embeddings, and predicts token-wise
relevance through lightweight fusion attention and an MLP. Its input features
are detached so that the relevance-prediction loss does not directly update
the backbone through the gate-input path. The resulting attention bias remains
differentiable and is optimized jointly with the WAM.

\subsection{Training, Compute, and Action-Only Inference}

\begin{center}
\small
\begin{tabularx}{\linewidth}{@{}lY lY@{}}
\toprule
\textbf{Training item} & \textbf{Value}
& \textbf{Training item} & \textbf{Value} \\
\midrule
Optimizer & AdamW
& Learning rate & $10^{-4}$ \\
Schedule & Cosine
& Warmup & 5\% of training \\
Weight decay & $10^{-2}$
& Gradient clipping & 1.0 \\
Batch / GPU & 1
& Gradient accumulation & 16 \\
Global batch & $16\times$ number of GPUs
& Training length & 10 epochs \\
Precision & BF16
& Diffusion train steps & 1000 \\
Input resolution & $224\times448$
& Inference steps & 10 \\
$\lambda_{\mathrm{TD}}$ & 1.0
& $\lambda_{\mathrm{Track}}$ & 0.2 \\
$\lambda_{\mathrm{Route}}$ & 0.5
& Dice weight & 0.25 \\
Trainable modules & DiT + GateNet
& Validation split & 0.05 for single-task runs \\
\bottomrule
\end{tabularx}
\captionof{table}{Final optimization and training configuration.}
\label{tab:training_hparams}
\end{center}

\begin{center}
\begin{minipage}[t]{0.485\linewidth}
\vspace{0pt}
\centering
\small
\begin{tabularx}{\linewidth}{@{}lY@{}}
\toprule
\textbf{Compute item} & \textbf{Configuration} \\
\midrule
GPU & $8\times$ A100 80GB \\
CUDA / PyTorch & cu128 / 2.7.1+cu128 \\
Python & $\ge 3.10$ \\
Training stack & Accelerate, DeepSpeed 0.18.5, Hydra 1.3.2 \\
Wan2.2 & HuggingFace TI2V-5B \\
CoTracker3 & Offline checkpoint \\
SAM & ViT-B checkpoint \\
\bottomrule
\end{tabularx}
\captionof{table}{Hardware and software environment.}
\label{tab:compute}
\end{minipage}
\hfill
\begin{minipage}[t]{0.485\linewidth}
\vspace{0pt}
\centering
\small
\begin{tabularx}{\linewidth}{@{}lY@{}}
\toprule
\textbf{Inference item} & \textbf{Configuration} \\
\midrule
DynaRoute evaluation & Once at $\tau_{\mathrm{init}}=1$ \\
Prefill input & Clean observation + noisy future/action slots \\
Visual cache & Routed visual key--value cache \\
Video branch & Disabled after prefill \\
Action denoising & 10 steps \\
Inference precision & FP/BF16 on GPU \\
Latency / memory & -- \\
\bottomrule
\end{tabularx}
\captionof{table}{Action-only inference configuration.}
\label{tab:action_only_inference}
\end{minipage}
\end{center}


\section{Additional Results}

Figure~\ref{fig:liberoplus_levels} shows performance as perturbation severity
increases from L1 to L5. DC-WAM maintains the clearest advantage at moderate
severity levels, while the most severe L5 shifts remain challenging for all
methods.

\begin{center}
\begin{minipage}[t]{0.47\linewidth}
\vspace{0pt}
\centering
\small
\begin{tabular}{@{}lccccc@{}}
\toprule
\textbf{Method} & \textbf{Spat.} & \textbf{Obj.} & \textbf{Goal} & \textbf{Long} & \textbf{Avg.} \\
\midrule
FastWAM$^\dagger$ & 98.2 & 100.0 & 97.0 & 95.2 & 97.6 \\
FastWAM-AC & 97.0 & 99.4 & 95.6 & 94.8 & 96.7 \\
\textbf{DC-WAM} & \textbf{98.8} & 99.4 & \textbf{98.0} & \textbf{96.2} & \textbf{98.1} \\
\bottomrule
\end{tabular}
\captionof{table}{Success rate (\%) on the four LIBERO suites. $\dagger$
denotes values reported in the original paper.}
\label{tab:libero_main}
\end{minipage}
\hfill
\begin{minipage}[t]{0.49\linewidth}
\vspace{0pt}
\centering
\includegraphics[width=\linewidth]{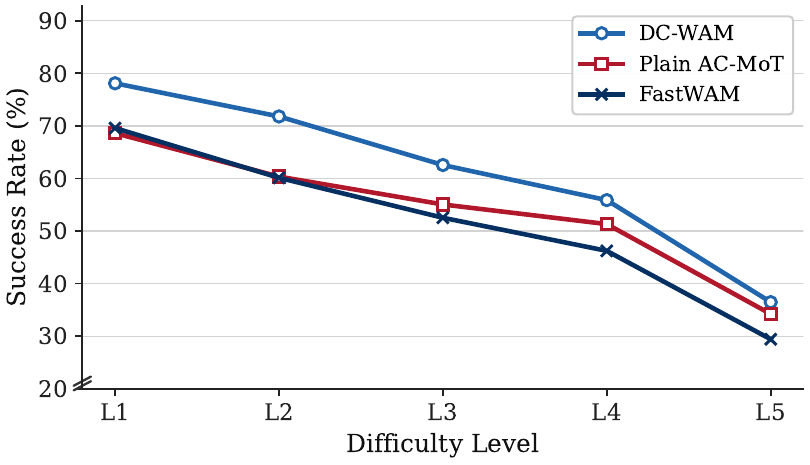}
\captionof{figure}{Success rate under progressive LIBERO-Plus perturbations
from L1 to L5.}
\label{fig:liberoplus_levels}
\end{minipage}
\end{center}

\subsection{Completion Time and Inference Efficiency}

\begin{center}
\begin{minipage}[t]{0.485\linewidth}
\vspace{0pt}
\centering
\small
\begin{tabular}{@{}lccc@{}}
\toprule
\textbf{Task} & \textbf{Clean} & \textbf{Light} & \textbf{Bg.} \\
\midrule
Stack-Bowl & -- & -- & -- \\
Pile-Plates & -- & -- & -- \\
Collect-Potato & -- & -- & -- \\
\bottomrule
\end{tabular}
\captionof{table}{Mean completion time for successful real-world trials.}
\label{tab:completion_time}
\end{minipage}
\hfill
\begin{minipage}[t]{0.485\linewidth}
\vspace{0pt}
\centering
\small
\begin{tabular}{@{}lcccc@{}}
\toprule
\textbf{Method} & \textbf{Video} & \textbf{Bias} & \textbf{Latency} & \textbf{SR} \\
 & \textbf{steps} & \textbf{time} & \textbf{(ms)} & \\
\midrule
Full joint & -- & -- & -- & 97.4 \\
Cache (ours) & -- & -- & -- & -- \\
\bottomrule
\end{tabular}
\captionof{table}{Inference efficiency at batch size one.}
\label{tab:efficiency}
\end{minipage}
\end{center}

\section{More Discussion on Dynaic Map}

\paragraph{View-dependent dynamic maps.}
Wrist-camera motion can make a large fraction of the image appear dynamic.
We therefore construct and normalize maps separately for each camera view and
use representative point sampling to concentrate supervision around the
gripper, manipulated objects, and contact regions.

\paragraph{Dependence on offline perception tools.}
DC-WAM uses off-the-shelf tracking and segmentation only to construct
training targets. Tracking errors, missed regions, severe occlusion, and large
camera motion can nevertheless degrade the supervision signal. The current
implementation is designed for RGB-based visual--action MoT architectures;
extending it to mobile manipulation, deformable objects, or substantially
different camera configurations may require revised target construction and
routing mechanisms.
